\documentclass[runningheads]{llncs}
\usepackage{graphicx}
\usepackage{comment}
\usepackage{amsmath,amssymb} 

\usepackage{color}
\usepackage{epsfig}
\usepackage{booktabs}       
\usepackage{cite}
\usepackage{multirow}
\usepackage{makecell}
\usepackage{tabularx}
\usepackage{arydshln}
\usepackage{subfigure}
\usepackage{balance}

\usepackage{amsmath,amsfonts,bm}

\def\eqref#1{equation~\ref{#1}}

\def\1{\bm{1}}

\newcommand{\test}{\mathcal{D_{\mathrm{test}}}}

\DeclareMathAlphabet{\mathsfit}{\encodingdefault}{\sfdefault}{m}{sl}
\SetMathAlphabet{\mathsfit}{bold}{\encodingdefault}{\sfdefault}{bx}{n}

\newcommand{\KL}{D_{\mathrm{KL}}}

\usepackage{mathtools}
\usepackage{algorithm}
\usepackage{algorithmic}

\begin{document}
\pagestyle{headings}
\mainmatter
\def\ECCVSubNumber{1035}  

\title{Learning to Learn with Variational Information Bottleneck for Domain Generalization} 

\titlerunning{Learning to Learn with Variational Information Bottleneck}
%
\author{Yingjun Du\inst{1}\orcidID{0000-0001-7537-6457} \and
Jun Xu\inst{2} \and
Huan Xiong\inst{4}  \and
Qiang Qiu\inst{5} \and
Xiantong Zhen\inst{1, 3} \and
Cees G. M. Snoek\inst{1} \and
Ling Shao\inst{3, 4}
}

\authorrunning{Yingjun Du. et al.}
%
\institute{AIM Lab, University of Amsterdam, The Netherlands \and
College of Computer Science, Nankai University, China\\ \and
Inception Institute of Artificial Intelligence, Abu Dhabi, UAE\\ \and 
Mohamed bin Zayed University of Artificial Intelligence, Abu Dhabi, UAE \and
Electrical and Computer Engineering, Duke University, USA \\
\email{ \{y.du, x.zhen, cgmsnoek\}@uva.nl, \ nankaimathxujun@gmail.com, \ huan.xiong@mbzuai.ac.ae,  qiang.qiu@duke.edu, \ ling.shao@ieee.org}
}

\maketitle
\begin{abstract}
Domain generalization models learn to generalize to previously unseen domains, but suffer from prediction uncertainty and domain shift. In this paper, we address both problems. We introduce a probabilistic meta-learning model for domain generalization, in which classifier parameters shared across domains are modeled as distributions. This enables better handling of prediction uncertainty on unseen domains. To deal with domain shift, we learn domain-invariant representations by the proposed principle of meta variational information bottleneck, we call MetaVIB. MetaVIB is derived from novel variational bounds of mutual information, by leveraging the meta-learning setting of domain generalization. Through episodic training, MetaVIB learns to gradually narrow domain gaps to establish domain-invariant representations, while simultaneously maximizing prediction accuracy. We conduct experiments on three benchmarks for cross-domain visual recognition. Comprehensive ablation studies validate the benefits of MetaVIB for domain generalization. The comparison results demonstrate our method outperforms previous approaches consistently.

\keywords{Meta Learning, Domain Generalization, Variational Inference, Information Bottleneck}
\end{abstract}

\section{Introduction}
This paper strives for domain generalization in image classification \cite{muandet2013domain,li2017deeper,vlcs,li2018deep}. The general challenge is to exploit the data variations of seen image domains with the aim to generalize well to unseen image domains. For example, by generalizing a \textit{chair} classifier trained on PASCAL VOC to LabelMe~\cite{vlcs}, or by generalizing an  \textit{elephant} classifier trained on photo's to sketches~\cite{li2017deeper}. Domain generalization models typically suffer from two problems. First, since data from unseen domains is inaccessible during the learning stage, we do not know their statistical data distribution. This causes uncertainty in the predictions made on the unseen domains. Second, data from different domains usually follows distinct distributions with great discrepancy, resulting in domain shift from seen to unseen domains. 
Domain shift has been extensively researched in domain generalization, mostly by learning feature representations that are invariant across domains~\cite{muandet2013domain,erfani2016robust,ghifary2015domain,xie2017controllable,li2018domain,li2018learning,li2019feature}. Meta-learning~\cite{schmidhuber1997shifting,thrun2012learning} that learns to generalize across tasks has been introduced to domain generalization by Li et al.~\cite{li2018learning} showing its great effectiveness in learning to generalize across domains~\cite{li2018learning,balaji2018metareg,li2019feature}. To the best of our knowledge, none of these existing meta-learning methods deal with the prediction uncertainty on unseen domains.

In this paper, we address the two major domain generalization challenges jointly by one single probabilistic model under the meta-learning framework.
We model parameters of classifiers shared across domains as probabilistic distributions that we infer from the data of the seen domains. The probabilistic modeling enables us to better handle the prediction uncertainty on previously unseen domains \cite{finn2018,gordon2018meta}. To handle domain shift, we take inspiration from the information bottleneck (IB) theory \cite{tishby2000ib,alemi2016deep,amjad2019learning} which learns robust representations to enhance  generalization. IB encodes the input into compressed intermediate representations that maximize target prediction. It offers a promising technique to learn domain-invariant representations, but to the best of our knowledge has not yet been explored for domain generalization under the meta-learning framework. We propose the principle of meta variational information bottleneck (MetaVIB) for the optimization of the model. We derive MetaVIB from the variational bounds of mutual information by leveraging the meta-learning setting, and incorporate it as a data-driven regularizer into the optimization objective. 
    The parameters of all classifiers and the network are jointly optimized during the meta-training stage and applied to the unseen domain in the meta-test stage. By episodic training, MetaVIB enables the network to learn to gradually close the gaps between domains to achieve domain-invariant representations that alleviate domain shift, while simultaneously being able to obtain accurate predictions.

We conduct extensive experiments on three benchmarks for cross-domain visual recognition. The ablation studies demonstrate the benefits of MetaVIB in the probabilistic framework for domain generalization. The comparison with state-of-the-art methods, shows that our method consistently delivers the best performance on all tasks, surpassing previous methods based on both regular learning and meta-learning.

\section{Related Work}
In this section, we review related work on domain generalization, information bottleneck and meta-learning.

\textbf{Domain generalization} has been a longstanding challenge in computer vision~\cite{li2017deeper,li2018domain,li2018learning} and machine learning \cite{blanchard2011generalizing,muandet2013domain},but recently regained increased research interest~\cite{shankar2018generalizing,carlucci2019domain,li2019feature,dou2019domain,balaji2018metareg}.
Learning domain-variant feature representation has been one of the main topics of focus in domain generalization~\cite{muandet2013domain,xie2017controllable,erfani2016robust,ghifary2015domain,li2018deep,li2018domain}. The core idea is to learn a model that generates invariant representations for the source domains, without over-fitting, which generalizes to unseen target domains. 
Muandet et al. \cite{muandet2013domain} propose a kernel-based optimization algorithm to learn an invariant transformation.
Li et al.~\cite{li2018domain} introduce adversarial auto-encoders to learn a generalized latent feature representation across domains. 
Their maximum mean discrepancy measure\textbf{} aligns distributions to learn universal representations to be independent of domains. %
We explore the domain discrepancy to learn invariant representations through the lens of mutual information \cite{tishby2000ib}.

\textbf{Information bottleneck (IB)}~\cite{tishby2000ib} provides an information-theoretic principle of encoding the input data into a compressed representation that maximizes target prediction. This is achieved by minimizing the mutual information $I(Z;X)$ between the input variable $X$ and its latent representation $Z$, while maximizing the mutual information $I(Z;Y)$ between the output variable $Y$ and the latent representation $Z$. To be more precise, the IB principle is to maximize the objective function:
\begin{equation}
\mathcal{L}_{\mathrm{IB}}(\boldsymbol{\theta})=I(Z;Y|\boldsymbol{\theta}) - \beta I(Z; X|\boldsymbol{\theta}),
\label{eqn:e6}
\end{equation}
where $\beta\in[0,\ 1]$ is the hyperparameter that controls the size of the information bottleneck, and $\boldsymbol{\theta}$ are the corresponding model parameters.

The IB principle has recently been introduced for theoretical understanding and analysis of deep neural networks~\cite{Tishby2015dlib,shwartz2017opening,amjad2019learning,peng2018variational,kolchinsky2018caveats}. The authors optimize the networks with an iterative Blahut-Arimoto algorithm, which is infeasible in practical systems. Alemi et al.~\cite{alemi2016deep} developed a variational approximation to the IB objective by leveraging variational inference, which allows the IB model to be parameterized with neural networks. Amjad et al. ~\cite{amjad2019learning} investigated training deep neural networks (DNN) for classification based on minimization of the IB functional.
It is shown that for deterministic DNNs, the optimization can be ill-posed. This is because the IB functional can be infinite or not admitting gradient descent since it is piece-wise constant. 
The possible remedy indicated in their work is to train stochastic DNNs with the IB principle.

\textbf{Meta-learning}, or learning to learn, endows models with the capacity to efficiently learn new tasks by acquiring common knowledge through experiencing a set of related tasks. It has been explored in several directions, e.g., by learning a meta learner on diverse tasks to adapt the parameters of the base learner on a specific task~\cite{Vinyals2016,snell2017prototypical,finn2017model,sung2018learning,satorras2018few, zhen2020learning}, learning to optimize the parameters of deep neural networks~\cite{Schmidhuber1992,bertinetto2016learning,munkhdalai2017meta}, and learning to learn the gradient optimization process by recurrent neural networks~\cite{andrychowicz2016learning,ravi2017optimization}, \textsl{etc}. 
A representative meta-learning algorithm is the model agnostic meta-learning (MAML), which learns the models to be able to adapt to similar tasks with only a few  gradient descent updates. Li et al.~\cite{li2018learning} introduced the idea of MAML~\cite{finn2017model} to domain generalization. They train models with generalization ability to unseen domains by leveraging the meta-learning setting. MetaReg~\cite{balaji2018metareg} addresses the domain shifts by leveraging the insights from meta-learning~\cite{vilalta2002perspective}. They learn a meta regularizer to achieve the generalization from source to unseen target domains. Li et al. \cite{li2019feature} proposed a meta-learning approach based on a feature-critic network, in which an auxiliary loss is introduced to improve generalization ability. Dou et al. \cite{dou2019domain} adopt a gradient-based model-agnostic learning algorithm to deal with domain shift for domain generalization. Two complementary losses are introduced for regularization of semantic features. The success of those works has indicated the effectiveness of meta-learning in domain generalization. 
Probabilistic meta-learning has also been developed in few-shot learning to handle uncertainty~\cite{finn2018probabilistic,gordon2018meta}, which has not been explored for domain generalization. 

In this work, we introduce a probabilistic meta-learning model for domain generalization which enables better handling of prediction uncertainty on unseen domains. We introduce the IB principle for domain-invariant representation learning by a stochastic deep neural network. We derive a new variational approximation to the IB principle under the meta-learning framework, resulting in the meta variational information bottleneck (MetaVIB) principle for domain generalization. We adapt the episodic training strategy in meta-learning by using the meta-train and meta-test splits of the source domains in each mini-batch for stochastic optimization.

\section{Method}

We describe the meta-learning setting for domain generalization. Following the setting in recent domain generalization by meta-learning~\cite{li2018learning,balaji2018metareg,li2019feature}, we divide a dataset into the Source domains $\mathcal{S}$ used for train and the Target domains $\mathcal{T}$ held-out for test. In the train phase, data in the source domains $\mathcal{S}$ is episodically divided into sets of meta-train $\mathcal{D}^{s}$ and  meta-test $\mathcal{D}^{t}$ domains. We train the model by optimizing over the prediction errors on meta-test $\mathcal{D}^{t}$ domains. In the test phase, the learned model is applied to the target domains $\mathcal{T}$ for performance evaluation. The training phase incorporates the idea of meta-learning which induces a higher level of learning by the split of meta-train and mete-test domains, rather than training on all source domains~\cite{Li_2019_ICCV}. This episodic meta-learning process mimics the generalization from seen to previously unseen domains.

\subsection{Probabilistic Modeling}
We start with the probabilistic formulation of the domain generalization, based on which we develop the probabilistic model under the meta-learning framework. We consider the general estimation problem of conditionally predictive likelihood in the meta-test domain $\mathcal{D}^{t}$:
\begin{equation}
\max \underset{p(\mathbf{x}^t,\mathbf{y}^t)}{\mathbb{E}}[\log \int p(\mathbf{y}^t|\psi,\mathbf{x}^t)p(\psi|\mathbf{x}^t)\mathrm{d}\psi],
\label{eqn:e1}
\end{equation}
where $(\mathbf{x}^t,\mathbf{y}^t)$ is the sample of paired input and label drawn from data distribution $p(\mathbf{x}^t,\mathbf{y}^t)$ in meta-test domain, $p(\mathbf{y}^t|\psi,\mathbf{x}^t)$ is the conditionally predictive distribution, $\psi$ is the parameter set of the classifier. Note that we treat $\psi$ as a stochastic variable that depends on the input $\mathbf{x}^t$ and the optimization of (\ref{eqn:e1}) is with respect to the parameters of probabilities.

\begin{figure*}[t]
\begin{center}
\includegraphics[width=.85\textwidth]{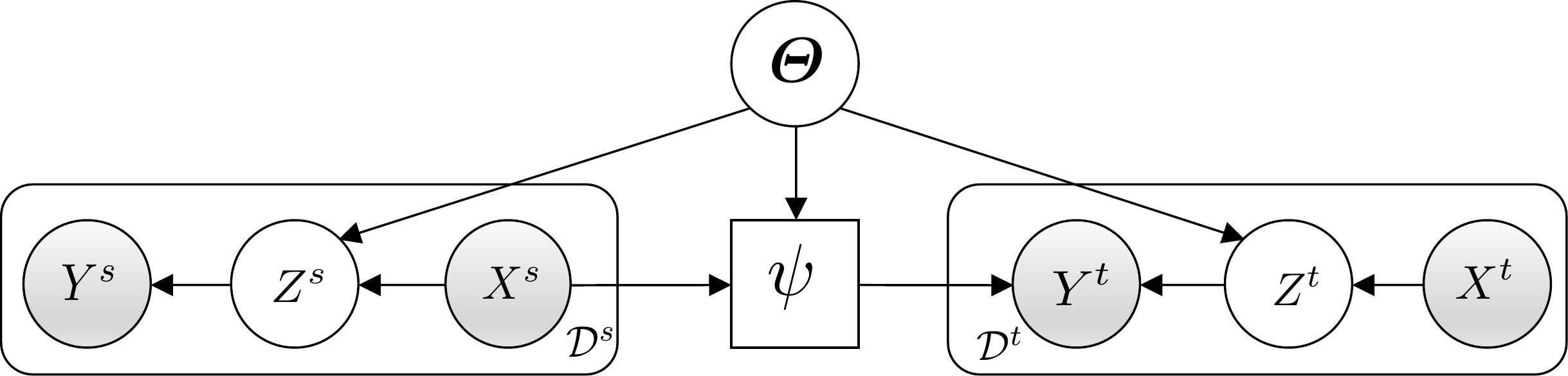}
\caption{\textbf{Computational graph of the probabilistic meta-learning model for domain generalization.} \boldsymbol{$\Theta$} encloses the global model parameters and $\psi$ contains the parameters of classifiers shared across domains. 
\boldsymbol{$\Theta$} and $\psi$ are jointly optimized in the train phase on the source domains. In each episode, the source domain is divided into a meta-train ($\mathcal{D}^s$) and meta-test ($\mathcal{D}^t$) domain. $\psi$ is produced by $\mathcal{D}^s$ and applied to $\mathcal{D}^t$. In the test phase, the model (\boldsymbol{$\Theta$}) generates representations of data in the target domains and the classifier ($\psi$) predicts of data from the source domain.}
\label{fig:dgm}
\end{center}
\end{figure*}

In this work, we parameterize the model by deep neural networks. From the information-theoretic point of view \cite{alemi2016deep}, we regard the feature representation from the neural network as a stochastic variable $\mathbf{z}^{t}$, which is the latent encoding of the input $\mathbf{x}^{t}$.
In domain generalization, it is commonly assumed that the label space is shared across the source and target domains. By leveraging the meta-learning setting, we propose to use data $D^s$ from the meta-train domains to estimate the parameters of the classifier by replacing $p(\psi|\textbf{x}^t)$ with $q(\psi|D^s)$, which is applied to the meta-test domain. By incorporating the latent variable $\textbf{z}$ into (\ref{eqn:e1}), we obtain the following maximum conditionally predictive likelihood estimation,
\begin{equation}
\max \underset{p(\mathbf{x}^t,\mathbf{y}^t)}{\mathbb{E}} [\log\int p(\mathbf{y}^t|\psi,\mathbf{z}^t)p(\mathbf{z}^t|\mathbf{x}^t)q(\psi|D^s)\mathrm{d}\mathbf{z}\mathrm{d}\psi].
\label{eqn:e4}
\end{equation}
This establishes a probabilistic latent model which can be represented in a computational graph as shown in Fig.~\ref{fig:dgm}, and the corresponding conditional joint distribution is defined as:
\begin{equation}
\begin{aligned}
p(Y^{t},Z^t,\psi&|X^{t},D^s,\boldsymbol{\Theta})= p(\psi|D^s;\boldsymbol{\Theta}) \prod_{n=1}^{N}
p(\mathbf{y}_n^{t}|\psi,\mathbf{z}^t_n)p(\mathbf{z}^t_n|\mathbf{x}_n^{t};\boldsymbol{\Theta}),
\end{aligned}
\label{eqn:e5}
\end{equation}
where $\boldsymbol{\Theta}$ denotes the model parameters, $D^t=\{X^t,Y^t\}=\{\mathbf{x}_n^{t},\mathbf{y}_n^{t}\}^N_{n=1}$, $D^s=\{X^s,Y^s\}=\{\mathbf{x}_m^{s},\mathbf{y}_m^{s}\}^M_{m=1}$, and $N$ ($M$) are the number of samples in the meta-test (meta-train) domains. It is possible to directly employ (\ref{eqn:e4}) as the optimization objective using the techniques of amortized inference~\cite{kingma2013auto,rezende2014stochastic}. However, the learned representations $\mathbf{z}$ would not be domain invariant, which is desired for domain generalization. To achieve domain-invariant representations, we resort to the information bottleneck (IB) principle~\cite{Tishby2015dlib,alemi2016deep}, which will be incorporated into the objective as a regularizer for joint optimization.

\subsection{Meta Variational Information Bottleneck}
We introduce the IB principle to learn domain-invariant representations under the meta-learning framework. We impose the information bottleneck on the feature representations to control the information flow in deep neural networks. This should largely remove domain related information while letting through the information that maximizes prediction of labels on the meta-test domain. 

We derive new variational bounds of mutual information by leveraging the setting of meta-learning for domain generalization. This gives rise to a meta version of variational information bottleneck, which we call MetaVIB in contrast to its original form in a standard learning framework~\cite{alemi2016deep}. To avoid confusion, we omit the superscript $t$ for the meta-test domain in this subsection. 

Let the random variables $X$, $Y$, and $Z$ denote the input, output, and the intermediate feature representation in the deep neural network, which encodes $X$. The mutual information $I(Z;Y)$ between the latent encoding $Z$ of data $X$ and its output label $Y$ is defined as follows:
\begin{equation}
\begin{aligned}
I(Z;Y)
&=\int p(\mathbf{y},\mathbf{z})\log\frac{p(\mathbf{y},\mathbf{z})}{p(\mathbf{y})p(\mathbf{z})}\mathrm{d}\mathbf{y}\mathrm{d}\mathbf{z}
=\int p(\mathbf{y},\mathbf{z})\log\frac{p(\mathbf{y}|\mathbf{z})}{p(\mathbf{y})}\mathrm{d}\mathbf{y}\mathrm{d}\mathbf{z}.
\end{aligned}
\label{eqn:e7}
\end{equation}
Since $p(\mathbf{y}|\mathbf{z})$ is intractable, we introduce $q(\mathbf{y}|\mathbf{z},\psi)$ to be a variational approximation of $p(\mathbf{y}|\mathbf{z})$, where conditioning on the classifier parameter $\psi$ is indicated by (\ref{eqn:e5}), and the prior distribution of $\psi$ is denoted as $p(\psi)$. Then we have:
\begin{equation}
    D_{\mathrm{KL}}[p(\mathbf{y}|\mathbf{z})||q(\mathbf{y}|\mathbf{z}, \psi)] = \int p(\mathbf{y}|\mathbf{z}) \log \frac{p(\mathbf{y}|\mathbf{z})}{q(\mathbf{y}|\mathbf{z},\psi)}\mathrm{d}\mathbf{y} \geq 0
    ,
\label{eqn:e8}
\end{equation}
which leads to 
\begin{equation}
\begin{aligned}
I(Z;Y)&\geq\int p(\mathbf{y},\mathbf{z})\log q(\mathbf{y}|\mathbf{z}, \psi)\mathrm{d}\mathbf{y}\mathrm{d}\mathbf{z} + H(Y),
\end{aligned}
\label{eqn:e10}
\end{equation}
where $H(Y)=-\int p(\mathbf{y})\log p(\mathbf{y})d\mathbf{y}$ is the entropy of $Y$.\
Taking expectation values of both sides with respect to $\psi\sim p(\psi)$, we have
\begin{equation}
\begin{aligned}
I(Z;Y)-H(Y)& \geq \mathbb{E}_{\psi\sim p(\psi)} \int p(\mathbf{y},\mathbf{z})\log q(\mathbf{y}|\mathbf{z}, \psi)\mathrm{d}\mathbf{y}\mathrm{d}\mathbf{z}
\\&
=  \int p(\psi) p(\mathbf{y},\mathbf{z}) \log q(\mathbf{y}|\mathbf{z}, \psi)\mathrm{d}\mathbf{y}\mathrm{d}\mathbf{z}\mathrm{d}\psi.
\label{eqn:e11}
\end{aligned}
\end{equation}
Note that the entropy $H(Y)$ is independent of our optimization procedure and can thus be ignored.\ By replacing the prior $p(\psi)$ with a meta prior $q(\psi|D^s)$ conditioned on data $D^s$ from the meta-train domains, leveraging the fact that $p(\mathbf{y},\mathbf{z})= \int p(\mathbf{y},\mathbf{z}|\mathbf{x})p(\mathbf{x}) \mathrm{d}\mathbf{x} = \int p(\mathbf{y}|\mathbf{x})p(\mathbf{z}|\mathbf{x})p(\mathbf{x}) \mathrm{d}\mathbf{x}$, and ignoring the $H(Y)$ term, we obtain a new variational lower bound:
\begin{equation}
\begin{split}
&I(Z;Y)\geq
\int p(\mathbf{x})p(\mathbf{y}|\mathbf{x})p(\mathbf{z}|\mathbf{x})q(\psi|D^s) \log q(\mathbf{y}|\psi,\mathbf{z})\mathrm{d}\mathbf{x}\mathrm{d}\mathbf{y}\mathrm{d}\mathbf{z}\mathrm{d}\psi,
\end{split}
\label{eqn:e12}
\end{equation}
which is tractable in general by approximation~\cite{alemi2016deep}.

Now we consider the second term $I(Z;X)$, which can be written as follows:
\begin{equation}
\label{eqn:e13}
I(Z;X)=\int p(\mathbf{x},\mathbf{z})\log\frac{p(\mathbf{z}|\mathbf{x})}{p(\mathbf{z})}\mathrm{d}\mathbf{x}\mathrm{d}\mathbf{z}.
\end{equation}
Instead of simply using an uninformative prior $p(\mathbf{z})$, we leverage the meta setting and introduce a meta prior $q(\mathbf{z}|D^s)$ as a variational approximation  to $p(\mathbf{z})$.\ Due to the fact that $D_{\mathrm{KL}}[p(Z)||q(Z|D^s)]>0$, we obtain the following upper bound:
\begin{align}
I(Z;X)\le\int p(\mathbf{x})p(\mathbf{z}|\mathbf{x})\log\frac{p(\mathbf{z}|\mathbf{x})}{q(\mathbf{z}|D^{s})}\mathrm{d}\mathbf{x}\mathrm{d}\mathbf{z}.
\label{eqn:encoder}
\end{align}

By combining the two bounds (\ref{eqn:e12}) and (\ref{eqn:encoder}), we establish the meta variational information bottleneck (MetaVIB) 
\begin{equation}
\begin{aligned}
\mathcal{L}_{\mathrm{IB}} \geq
&\int \, p(\mathbf{x}) p(\mathbf{y}|\mathbf{x})p(\mathbf{z}|\mathbf{x})p(\psi|D^s)\log q(\mathbf{y}|\mathbf{z},\psi)d\mathbf{x} \, d\mathbf{y} \, \mathrm{d}\mathbf{z}\mathrm{d}\psi\\
&-\beta
\int p(\mathbf{x})p(\mathbf{z}|\mathbf{x})\log\frac{p(\mathbf{z}|\mathbf{x})}{q(\mathbf{z}|D^{s})}\mathrm{d}\mathbf{x}\mathrm{d}\mathbf{z} = \mathcal{L}_{\mathrm{MetaVIB}}
\end{aligned}
\end{equation}
which extends the IB theory \cite{tishby2000ib} into the meta-learning scenario, offering a new principle of learning domain-invariant representations for domain generalization.

We follow \cite{alemi2016deep} to approximate 
$p(\mathbf{x},\mathbf{y})=p(\mathbf{x})p(\mathbf{y}|\mathbf{x})$ and $p(\mathbf{x})$ with empirical data distribution
$p(\mathbf{x},\mathbf{y})=\frac{1}{N}\sum_{n=1}^{N}\delta_{\mathbf{x}_{n}}(\mathbf{x})\delta_{\mathbf{y}_{n}}(\mathbf{y})$ and $p(\mathbf{x})=\frac{1}{N}\sum_{n=1}^{N}\delta_{\mathbf{x}_{n}}(\mathbf{x})$, where $N$ is the number of samples in the meta-test domain. This essentially regards the data points $(\textbf{x}_n,\textbf{y}_n)$ and $\textbf{x}_n$ as the samples drawn from the data distributions $p(\textbf{x},\textbf{y})$ and $p(\textbf{x})$, respectively.

Thus, the approximated lower bound $\tilde{\mathcal{L}}_{\mathrm{MetaVIB}}$ in practice can be written as:
\begin{equation}
\begin{aligned}
\tilde{\mathcal{L}}_{\mathrm{MetaVIB}}
&=
\frac{1}{N} \sum_{n=1}^N\int[ p(\mathbf{z}_n|\mathbf{x}_n)  p(\psi|D^s) \log q(\mathbf{y}_n|\mathbf{z}_n, \psi)\\&- \beta \, p(\mathbf{z}_n|\mathbf{x}_n) \log \frac{p(\mathbf{z}_n|\mathbf{x}_n)}{ q(\mathbf{z}_n|D^s)}]\mathrm{d}\mathbf{z}_n\mathrm{d}\psi.
\end{aligned}
\label{mvib}
\end{equation}

We use Monte Carlo sampling to draw samples from $p(\psi|D^s)$ for $\psi$ and from $p(\textbf{z}_n|\textbf{x}_n)$ for $\textbf{z}_n$ in the lower bound of MetaVIB in (\ref{mvib}). We attain the following objective function:
\begin{equation}
\begin{aligned}
\mathcal{L} = 
&
-\frac{1}{NC} \sum_{c=1}^C \sum_{n=1}^{N_c}\big(\frac{1}{L_\mathbf{z} L_{\psi}}\sum_{\ell_\mathbf{z}=1}^{L_\mathbf{z}}\sum_{\ell_{\psi}=1}^{L_\psi}\log q(\mathbf{y}_n|\mathbf{z}^{(\ell_{\mathbf{z}})}, \psi^{(\ell_{\psi})}_c) 
\\
&
+\beta \KL \left[ p(\mathbf{z}|\mathbf{x}_n)|| q(\mathbf{z}|D^s_c) \right]\big).
\end{aligned}
\label{objective}
\end{equation}
where $C$ is the number of classes and $D^s_c$ contains the samples from the $c$-th category in the meta-train domains. We amortize the posterior distribution $q(\psi|D^s_c)$ and the meta prior $q(\mathbf{z}_n|D^s_c)$ across classes, that is, the variational distribution of each class is inferred individually by the samples from its corresponding class $D^s_c$, which further alleviates the computational overhead. In addition, the KL term can be calculated in a closed form. Here, to enable back-propagation, we adopt the re-parameterization trick~\cite{kingma2013auto}, that is, 
\begin{equation}
\mathbf{z}^{(\ell_\mathbf{z})}_n=f(\mathbf{x}_n,\epsilon^{(\ell_\mathbf{z})}),~~ \epsilon^{(\ell_\mathbf{z})}\sim \mathcal{N}(0, I)
\end{equation}
and 
\begin{equation}
    \psi_c^{(\ell_{\psi})} =f(D^s_c,\epsilon^{(\ell_{\psi})}),~~ \epsilon^{(\ell_{\psi})} \sim \mathcal{N}(0, I)
\end{equation}
where $f(\cdot)$ is a deterministic function which is usually parameterized by a multiple layer perception (MLP) and $L_\mathbf{z}$ and $L_{\psi}$ are the number of samples for $\mathbf{z}_n$ and $\psi_c$, respectively.

Taking a closer look at the objective (\ref{objective}), we observe that the first term is the negative log predictive likelihood in the meta-test domain, where the label $\mathbf{y}_n$ of $\mathbf{x}_n$ is predicted from its latent encoding $\mathbf{z}_n$ and the classifier parameter $\psi$.\ Minimizing the first term guarantees maximal prediction accuracy. The second term is the KL divergence between distributions of latent encoding of the sample in the target domain and that estimated by the samples from the same category in the meta-train domains. It is the minimization of the KL term in (\ref{objective}) that enables the model to learn domain-invariant representations. This is in contrast to the regular IB principle \cite{Tishby2015dlib,alemi2016deep} which is to compress the input and does not necessarily result in domain-invariant representations.

\subsection{Learning with Stochastic Neural Networks}
We implement the proposed model by end-to-end learning with stochastic neural networks that are comprised of convolutional layers and fully-connected layers. The inference is parameterized by a feed-forward multiple layer perception (MLP). During the training phase, given $K$ domains, we randomly sample one domain as the meta-test domain, the remaining $K-1$ domains are used as the meta-train domains.\ Then we choose a batch of $M$ samples $\{(\mathbf{x}_{m}^{s},\mathbf{y}_{m}^{s})\}_{m=1}^{M}$ from the meta-train domain $\mathcal{D}^{s}$, and a batch of $N$ samples $\{(\mathbf{x}_{n}^{t},\mathbf{y}_{n}^{t})\}_{n=1}^{N}$ from the meta-test domain $\mathcal{D}^{t}$.\ Note that $M$ samples from meta-train domains cover all the $C$ classes. For each sample $\mathbf{x}_{m,c}^{s}$ of the $c$-th class, we first extract its features via $h_{\theta}(\mathbf{x}_{m,c}^{s})$, where $h_{\theta}(\cdot)$ is the feature extraction network and we use permutation-invariant instance-pooling operations to get the mean feature $\overline{\mathbf{h}}_c^{s}$ of samples in the $c$-th class.\ The mean feature $\overline{\mathbf{h}}_c^{s}$ will be fed into a small MLP network $g_{\phi_1}(\cdot)$ to calculate the mean $\boldsymbol{\mu}_{c}^{\psi}$ and variance $\boldsymbol{\sigma}_{c}^{\psi}$ of the weight vector distribution $\psi_c$ for $c$-th class, which is then used to sample the weight vector $\psi_c$ of this class by $\psi_c\sim\mathcal{N}(\boldsymbol{\mu}_{c}^{\psi}, \text{diag}(((\boldsymbol{\sigma}_{c}^{\psi})^2))$.\ The weight vectors $\{\psi_c\}_{c=1}^{C}$ of all $C$ classes are combined column by column to form a weight matrix $\psi=[\psi_1, \psi_2, ..., \psi_C]$.\ 

We calculate the parameters of the latent distribution, i.e., the mean $\boldsymbol{\mu}_{c}^{s}$ and variance $\boldsymbol{\sigma}_{c}^{s}$ of the $c$-th class in the meta-train domain by another small MLP network $g_{\phi_2}(\cdot)$.\ Then the parameter $\mathbf{z}_c$ is sampled from the distribution $\mathbf{z}_c\sim\mathcal{N}(\boldsymbol{\mu}_{c}^{s}, \text{diag}((\boldsymbol{\sigma}_{c}^{s})^2))$.\ For each sample $\mathbf{x}_{n,c}^{t}$ in the meta-test domain, we also calculate the mean $\boldsymbol{\mu}_{n,c}^{t}$ and variance $\boldsymbol{\sigma}_{n,c}^{t}$, of the distribution.\ Thus its latent coding vector $\mathbf{z}_{n,c}$ can be naturally sampled from $\mathbf{z}_{n,c}\sim\mathcal{N}(\boldsymbol{\mu}_{n,c}^{t}, \text{diag}(\mathbf{\sigma}_{n,c}^{t})^{2})$.\ Denote $\overline{\mathbf{h}}^s_c$ as the mean feature of all the samples of the $c$-th class from the meta-train domains, i.e., $\overline{\mathbf{h}}_c^s=\frac{1}{M_c}\sum\limits_{m=1}^{M_c}\mathbf{x}_{m,c}^{s}$. We provide the detailed step-by-step algorithm of the proposed MetaVIB for training in the supplemental material. 

\section{Experiments}
\label{sec:exp}
We conduct our experiments on three benchmarks commonly used in domain generalization~\cite{li2018learning,balaji2018metareg,shankar2018generalizing,li2019feature}. We first provide ablation studies to gain insights into the properties and benefits of MetaVIB. Then we compare with previous methods based on both regular learning and meta-learning for domain generalization. We put more results in the supplementary material due to space limit.

\subsection{Datasets}

\textbf{VLCS}~\cite{vlcs} is a real-world dataset that contains four domains collected from \textbf{V}OC2007~\cite{everingham2010pascal}, \textbf{L}abelMe~\cite{russell2008labelme}, \textbf{C}altech-101~\cite{griffin2007caltech}, and \textbf{S}UN09~\cite{choi2010exploiting}. Images are from 5 classes, i.e., \textsl{bird}, \textsl{car}, \textsl{chair}, \textsl{dog}, \textsl{person}. 
The domain shift across those datasets makes VLCS a suitable benchmark for domain generalization.

\noindent
\textbf{PACS}~\cite{li2017deeper} contains 9991 images 
from 4 domains, i.e., \textbf{P}hoto, \textbf{A}rt painting, \textbf{C}artoon, and \textbf{S}ketch, which cover huge domain gaps. Images are from 7 object classes, i.e., \textsl{dog}, \textsl{elephant}, \textsl{giraffe}, \textsl{guitar}, \textsl{horse}, \textsl{house}, and \textsl{person}.

\noindent
\textbf{Rotated MNIST}~\cite{shankar2018generalizing} is a synthetic dataset consisting of 6 domains, each containing 1000 images of the 10 digits (i.e., $\{0, 1,..., 9\}$, 100 for each) randomly selected from the training set of MNIST~\cite{lecun1998gradient}, with 
6 rotation degrees: $0^{\circ},15^{\circ},30^{\circ},45^{\circ},60^{\circ}$, and $75^{\circ}$.

\subsection{Implementation Details}
\noindent
\textbf{Splits, Metrics and Backbone}
On all datasets, we follow the train-test splits suggested by~\cite{li2017deeper,li2018learning, balaji2018metareg}, and perform experiments with the ``leave-one-domain-out'' strategy: we take the samples from one domain as the target domain for testing, and the samples from the remaining domains as the source domain for training.\ We use the AlexNet~\cite{krizhevsky2012imagenet} pre-trained on ImageNet and fine-tuned on the source domains of each dataset to perform testing on the target domain of that dataset.\ We use the average accuracy of all classes as the evaluation metric~\cite{ghifary2015domain}.\ To benchmark previous methods, we employ the pre-trained AlexNet~\cite{krizhevsky2012imagenet} on ImageNet as the backbone on VLCS and PACS.\ For Rotated MNIST we use a backbone network with two convolutions and one fully-connected layer.\  Even more implementation details about training stage, the feature extraction network and inference networks for different datasets are provided in the supplemental materials.


\subsection{Ablation Study}


\begin{table*}[t]
\setlength{\tabcolsep}{4pt}
\centering
    \caption{\textbf{Benefit of MetaVIB} under the probabilistic framework on VLCS}
\begin{tabular}{l l l l l l}
\toprule
 & \textbf{VOC2007} & \textbf{LabelMe} & \textbf{Caltech-101} & \textbf{SUN09} & \textbf{Mean} \\
\midrule
    AlexNet & 68.41  & 62.11  &  93.40  & 64.16  & 72.02 \\
    Baseline & 69.87{\scriptsize$\pm$0.63} &61.32{\scriptsize$\pm$0.27}    &  95.97{\scriptsize$\pm$0.43}&66.32{\scriptsize$\pm$0.25} & 73.37  \\
    VIB & 70.02{\scriptsize$\pm$0.52}  &62.17{\scriptsize$\pm$0.29}    & 95.93{\scriptsize$\pm$0.32} &67.93{\scriptsize$\pm$0.41} & 74.01  \\
    MetaVIB & 70.28{\scriptsize$\pm$0.71}  & 62.66{\scriptsize$\pm$0.35} & 97.37{\scriptsize$\pm$0.63} & 67.85{\scriptsize$\pm$0.17} & 74.54 \\
        \bottomrule
    \end{tabular}
 \label{tab:vlcs_metavib}
\end{table*}

\begin{table}[t]
\setlength{\tabcolsep}{6pt}
    \centering
    \caption{\textbf{Benefit of MetaVIB} under the probabilistic framework on PACS}
    \begin{tabular}{l l l l l l}
\toprule
& \textbf{Photo} & \textbf{Art painting} & \textbf{Cartoon}  & \textbf{Sketch} & \textbf{Mean} \\
\midrule
    AlexNet & 88.47   & 67.21  & 66.12    & 55.32  & 69.28 \\
    Baseline&90.32{\scriptsize$\pm$0.35} &68.12{\scriptsize$\pm$0.51}  & 70.25{\scriptsize$\pm$0.17}    &61.81{\scriptsize$\pm$0.26} & 72.63   \\
    VIB&90.17{\scriptsize$\pm$0.28} &69.93{\scriptsize$\pm$0.34}  & 71.01{\scriptsize$\pm$0.27}    &62.37{\scriptsize$\pm$0.42} & 73.37   \\
    MetaVIB & 91.93{\scriptsize$\pm$0.23} &71.94{\scriptsize$\pm$0.34}  & 73.17{\scriptsize$\pm$0.21}  &65.94{\scriptsize$\pm$0.24} & 75.74 \\
\bottomrule
    \end{tabular}
   \label{tab:pacs_metavib}
\end{table}

%
To study the benefit of the MetaVIB under the probabilistic framework for domain generalization, we compare with several alternative models on VLCS and PACS in  Tables~\ref{tab:vlcs_metavib} and \ref{tab:pacs_metavib}.

\subsubsection{Benefit of probabilistic modeling} To show the benefit of probabilistic modeling, we first consider AlexNet~\cite{krizhevsky2012imagenet} which is pre-trained on ImageNet, fine-tuned on the source domains and applied to the target domains. We define our \textit{Baseline} model as the probabilistic model that predicts parameter distributions of the classifiers, without regular VIB or MetaVIB. The probabilistic model outperforms the pre-trained AlexNet by $1.35\%$  and $3.35\%$  on the VLCS and PACS benchmarks. The results indicate that the classifiers learned by probabilistic modeling better generalize to the target domains. 
The further analysis of the prediction uncertainty of the probabilistic modeling is put in the supplemental materials.

\subsubsection{Benefit of MetaVIB}
%
We show the benefit of MetaVIB by comparing with the regular VIB \cite{alemi2016deep}, which is applied to the baseline model as a regularization in the optimization, and the Baseline model. We first establish the probabilistic model with the regular \textit{VIB} which performs better than the baseline (74.01\% - up 0.64\%) on VLCS and (73.37\% - up 0.74\%) on PACS. The VIB regularization term maximizes the mutual information between $Z$ and the target $Y$, which will encourage better prediction performance compared to the \textit{Baseline} model. However, our MetaVIB learns an even better domain-invariant representation, as it consistently outperforms VIB by up to $2.37\%$ on \textbf{PACS}~\cite{li2017deeper}. As indicated in the optimization objective in (\ref{objective}) minimizing the KL term makes the representations of samples in the meta-target domain to be close to the representations obtained by the samples of the same class from the meta-source domains. As a result, the learned model acquires the ability to generate domain-invariant representations by the episodic training. In contrast, the regular VIB is to simply compress the input with no explicit mechanism to narrow the gaps across domains. The obtained representations with regular VIB are not necessarily domain-invariant. Actually, there is no evident causal relation between compression and generalization as indicated in \cite{saxe2018information}.


\begin{figure*}[t]
\centering
\subfigure{
\begin{minipage}{0.23\textwidth}
\includegraphics[width=1\textwidth]{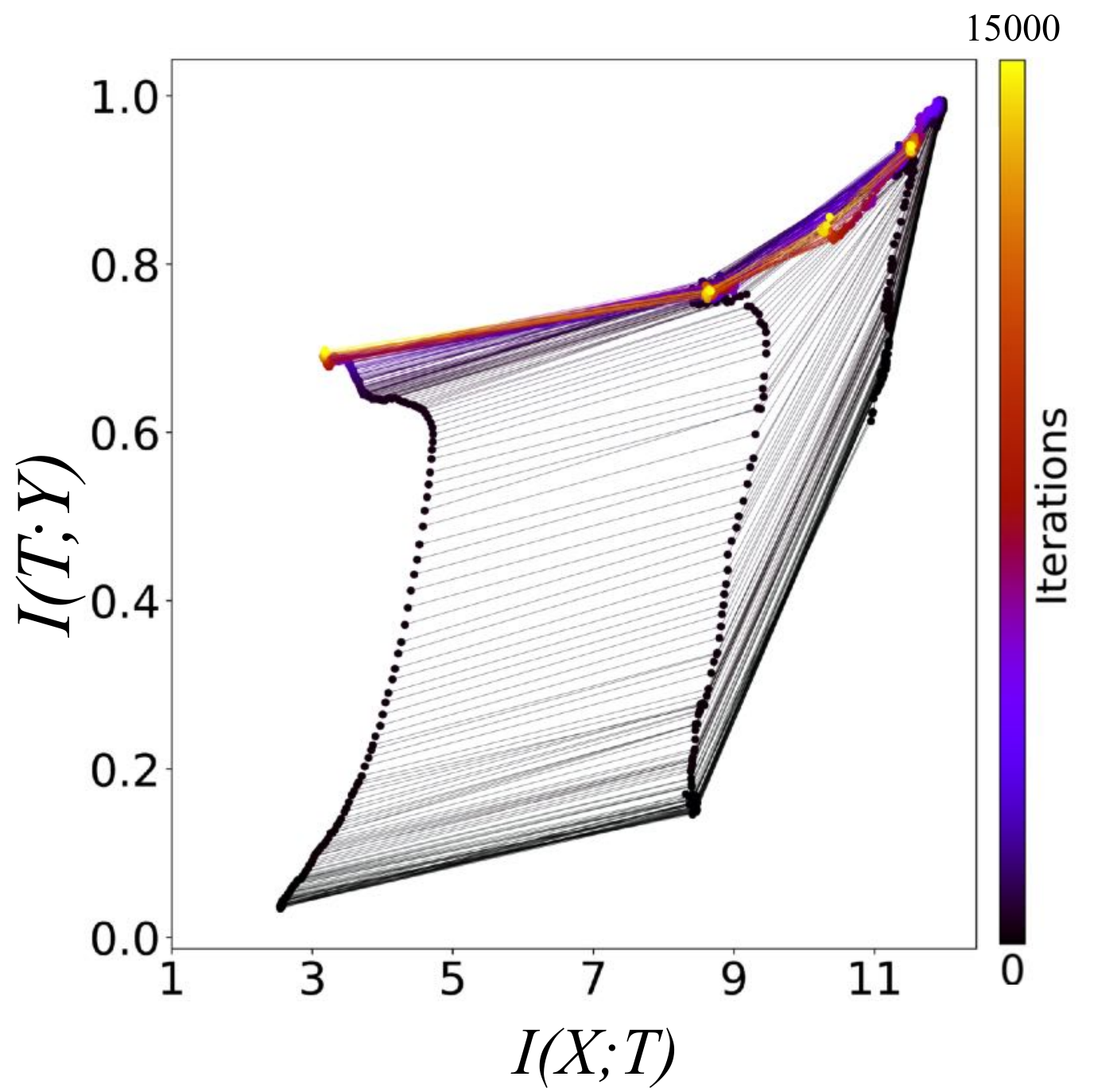}
\centering{(a) $\beta = 1$}
\end{minipage}
\begin{minipage}{0.23\textwidth}
\includegraphics[width=1\textwidth]{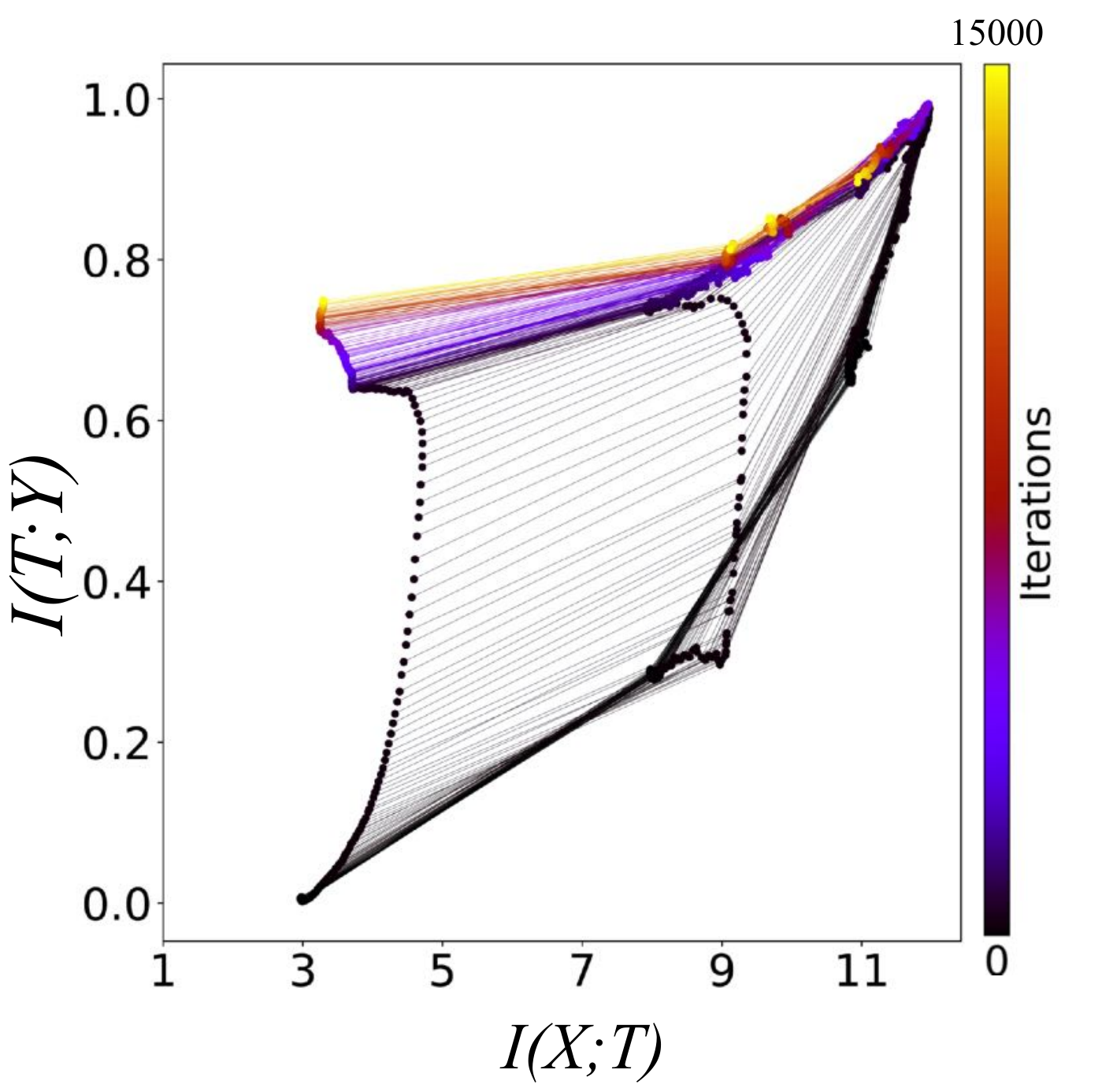}
\centering{(b) $\beta = 0.1$}
\end{minipage}
\begin{minipage}{0.23\textwidth}
\includegraphics[width=1\textwidth]{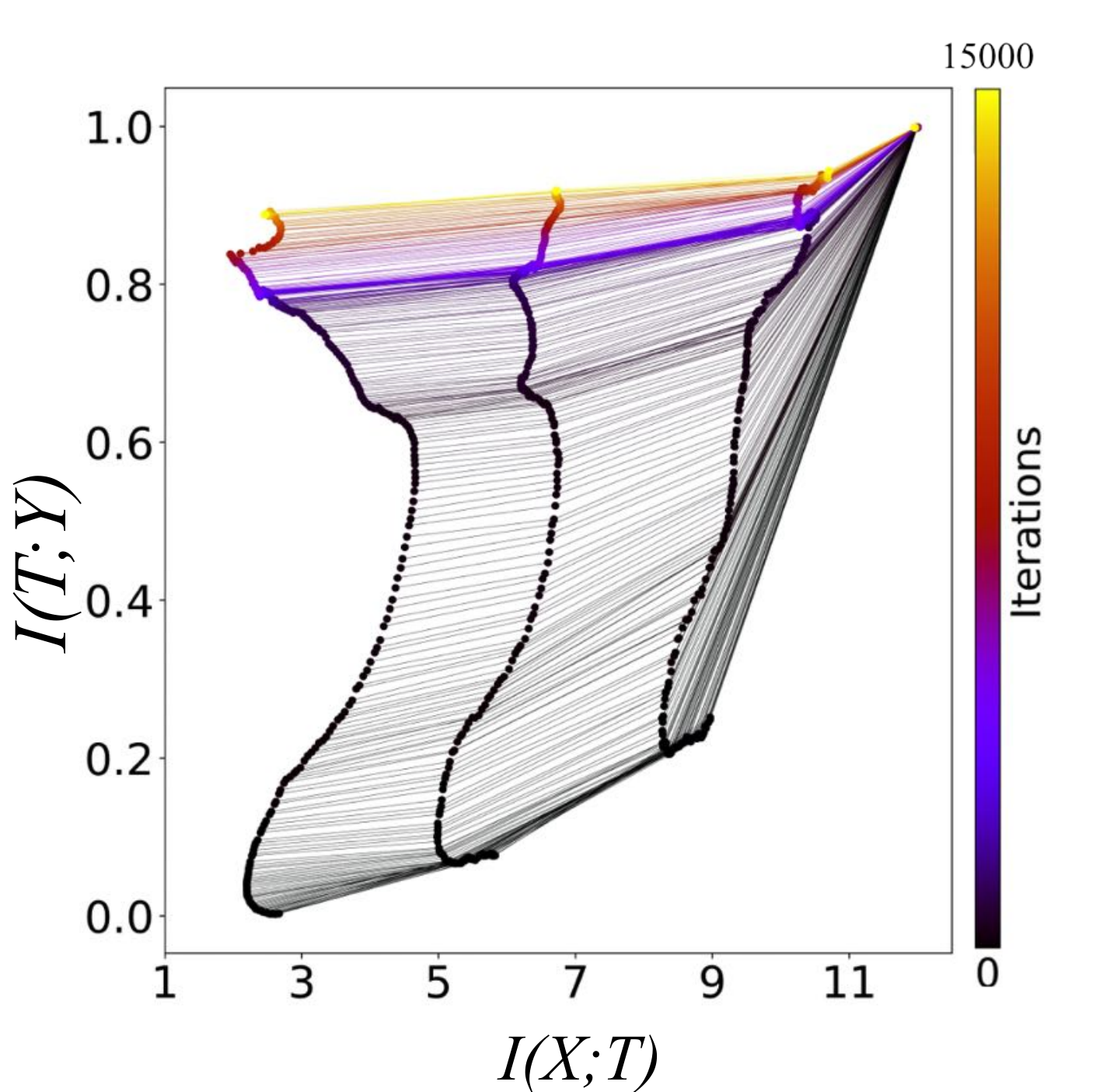}
\centering{(c) $\beta = 0.01$}
\end{minipage}
\begin{minipage}{0.23\textwidth}
\includegraphics[width=1\textwidth]{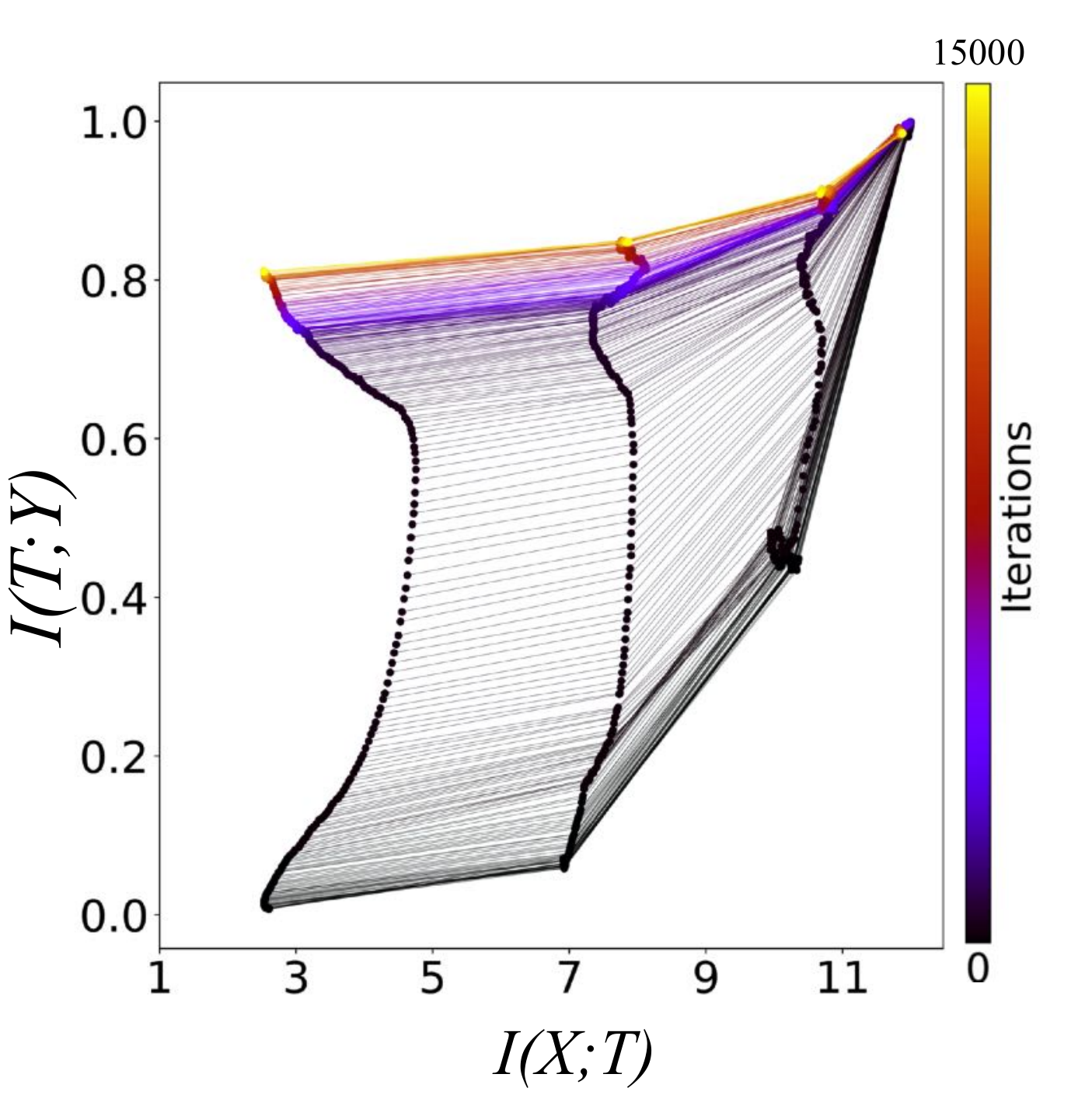}
\centering{(d) $\beta = 0.001$}
\end{minipage}
}
\caption{\textbf{Influence of information bottleneck size $\beta$} on domain generalization for  ``Sketch'' as the test domain on PACS.
$X$, $Y$, and $T$ denote input image, output target, and outputs per layer of the inference network that generates the latent encoding $Z$.\ The horizontal (vertical) axis plots mutual information between the features of each layer and the input (output).\ Each of the three layers of the inference network produces a curve in the information plane with the input layer at the far right and output layer at the far left. The color-scale denotes training iterations from $0$ to $15,000$. The mutual information of different layers in the same iteration are connected by fine lines. Compared to other values of $\beta$, for $\beta=0.01$, $I(T;Y)$ reaches the highest value, which explains the best performance.} 

\label{fig:beta_ip}
\end{figure*}

\begin{table*}[t]
\setlength{\tabcolsep}{6pt}
\centering
\caption{\textbf{Influence of information bottleneck size $\beta$} on domain generalization for PACS. MetaVIB obtains best results for $\beta=0.01$. We obtain similar results on other datasets, see supplemental material.} \label{tab:beta}
\begin{tabular}{lccccc}
\toprule
& \textbf{Photo} & \textbf{Art painting} & \textbf{Cartoon} & \textbf{Sketch} & \textbf{Mean} \\
\midrule
$\beta=1$& 89.05{\scriptsize$\pm$0.45} & 69.02{\scriptsize$\pm$0.41} & 71.13{\scriptsize$\pm$0.17}  & 58.87{\scriptsize$\pm$0.43} & 72.02 \\
$\beta=0.1$ & 90.51{\scriptsize$\pm$0.14} & 70.71{\scriptsize $\pm$0.28} & 70.78{\scriptsize$\pm$0.11} & 62.05{\scriptsize$\pm$0.26}& 73.51 \\
$\beta=0.01$&91.93{\scriptsize$\pm$0.23} &71.94{\scriptsize$\pm$0.34}  & 73.17{\scriptsize$\pm$0.21}  &65.94{\scriptsize$\pm$0.24} & 75.74  \\
$\beta=0.001$& 90.17{\scriptsize$\pm$0.25} & 70.07{\scriptsize$\pm$0.32}  & 71.75{\scriptsize$\pm$0.17}    & 63.90{\scriptsize$\pm$0.38} & 73.89 \\
\bottomrule
\end{tabular}
\end{table*}

\begin{figure*}[t]
\begin{center}
\subfigure{
\begin{minipage}{0.33\textwidth}
\includegraphics[width=1\textwidth]{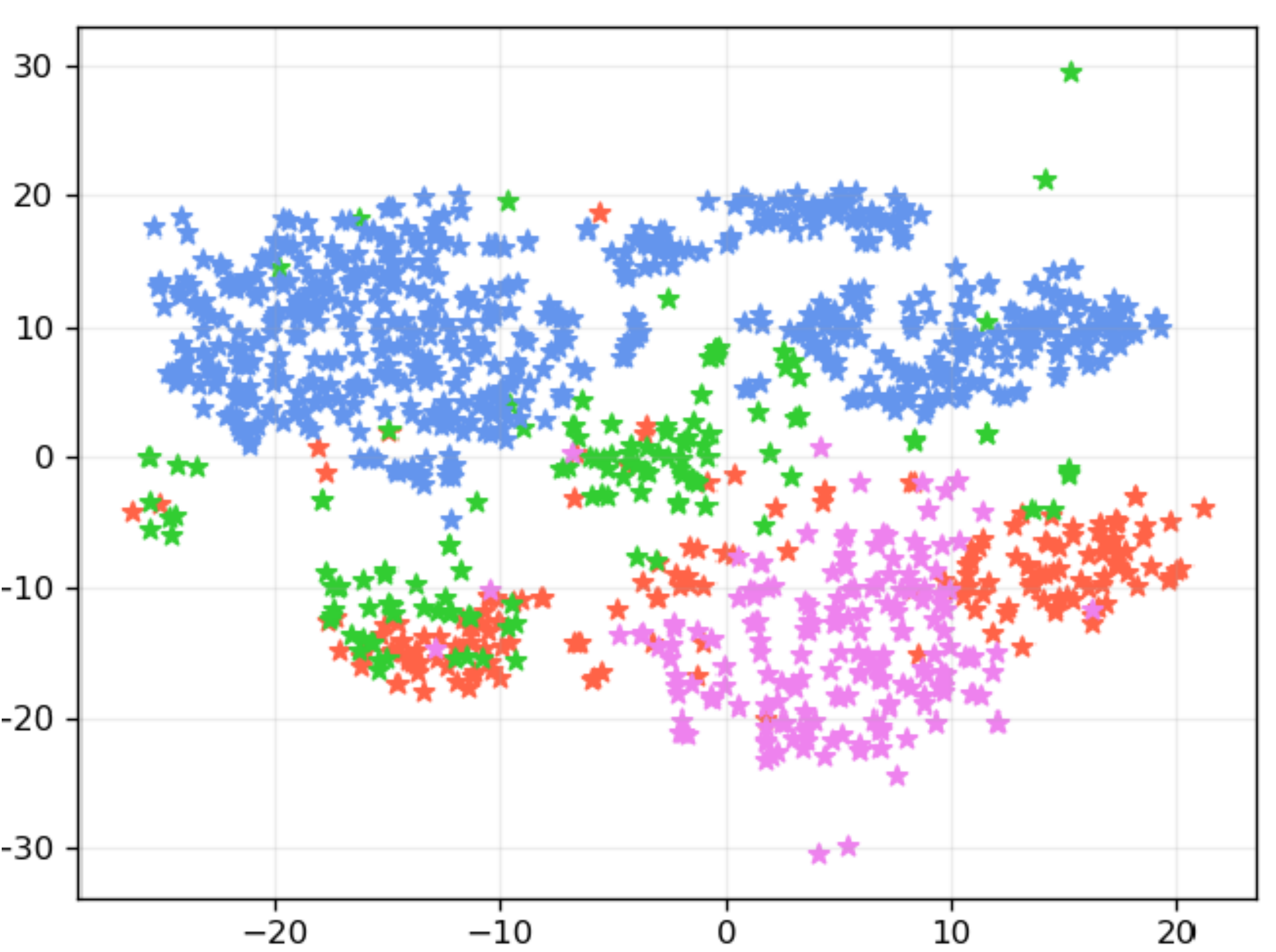}
\centering{(a) Pre-trained AlexNet}
\end{minipage}
\begin{minipage}{0.33\textwidth}
\includegraphics[width=1\textwidth]{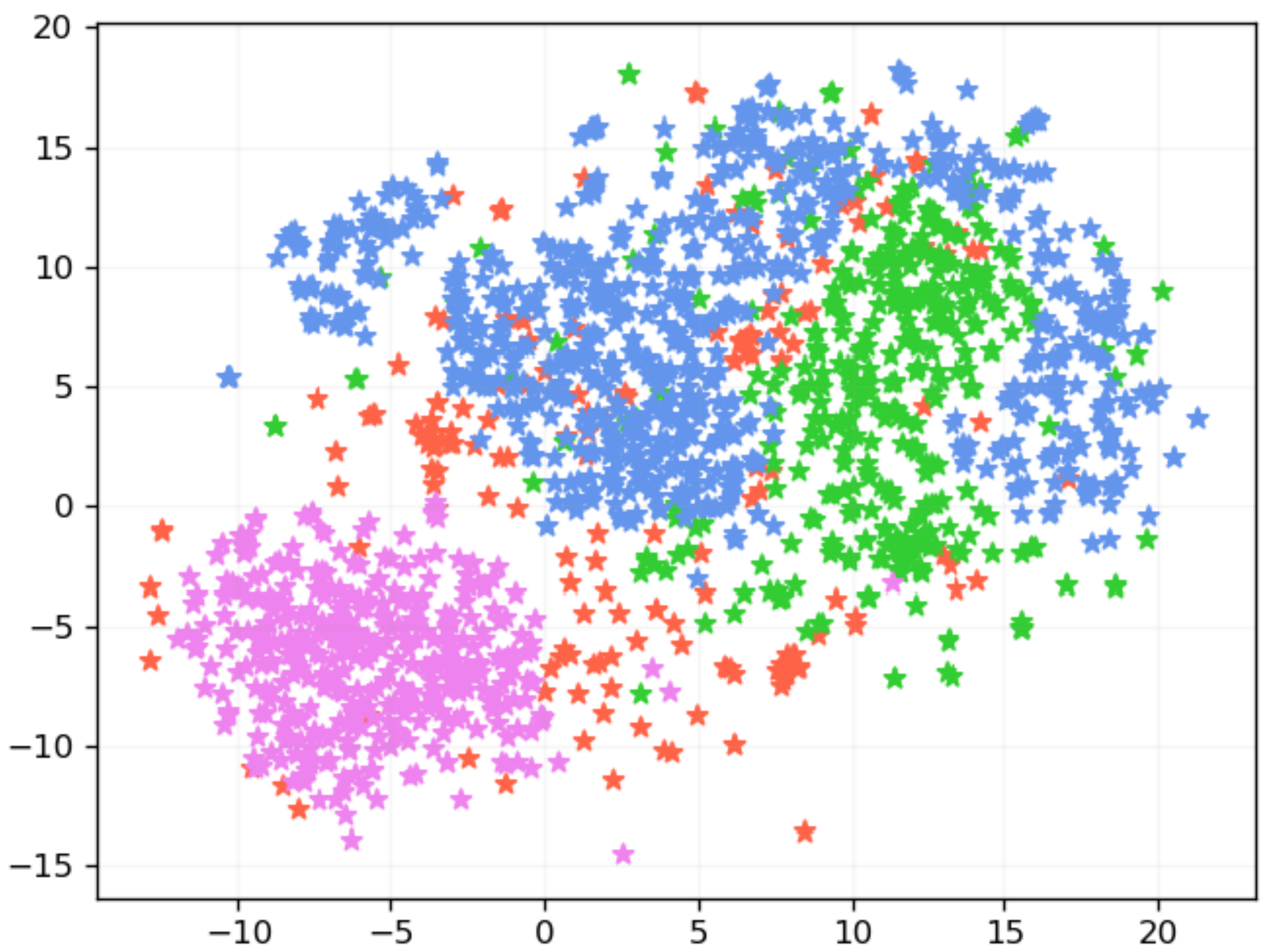}
\centering{(b) VIB}
\end{minipage}
\begin{minipage}{0.33\textwidth}
\includegraphics[width=1\textwidth]{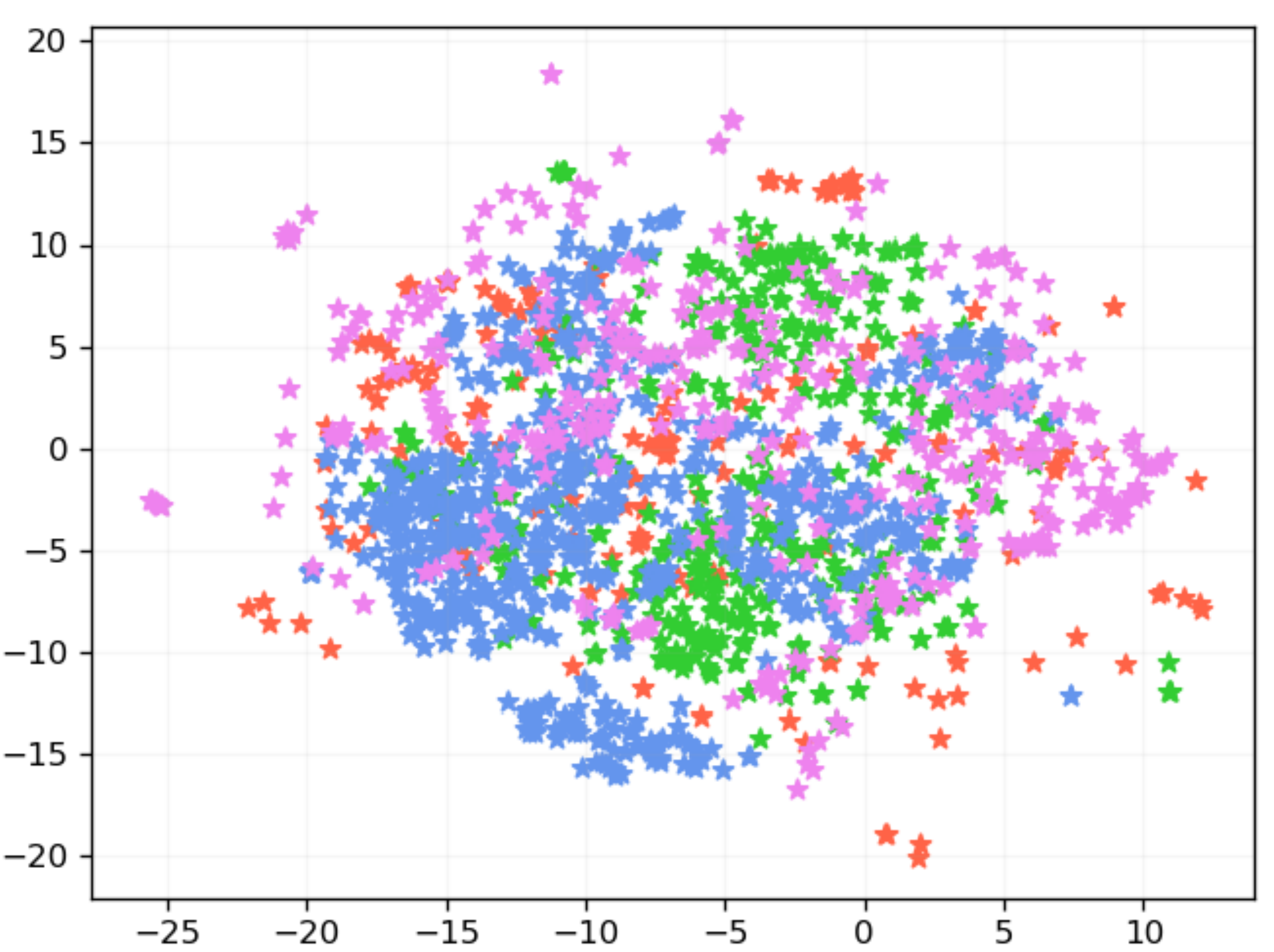}
\centering{(c) MetaVIB}
\end{minipage}
}
\subfigure{
\begin{minipage}{0.33\textwidth}
\includegraphics[width=1\textwidth]{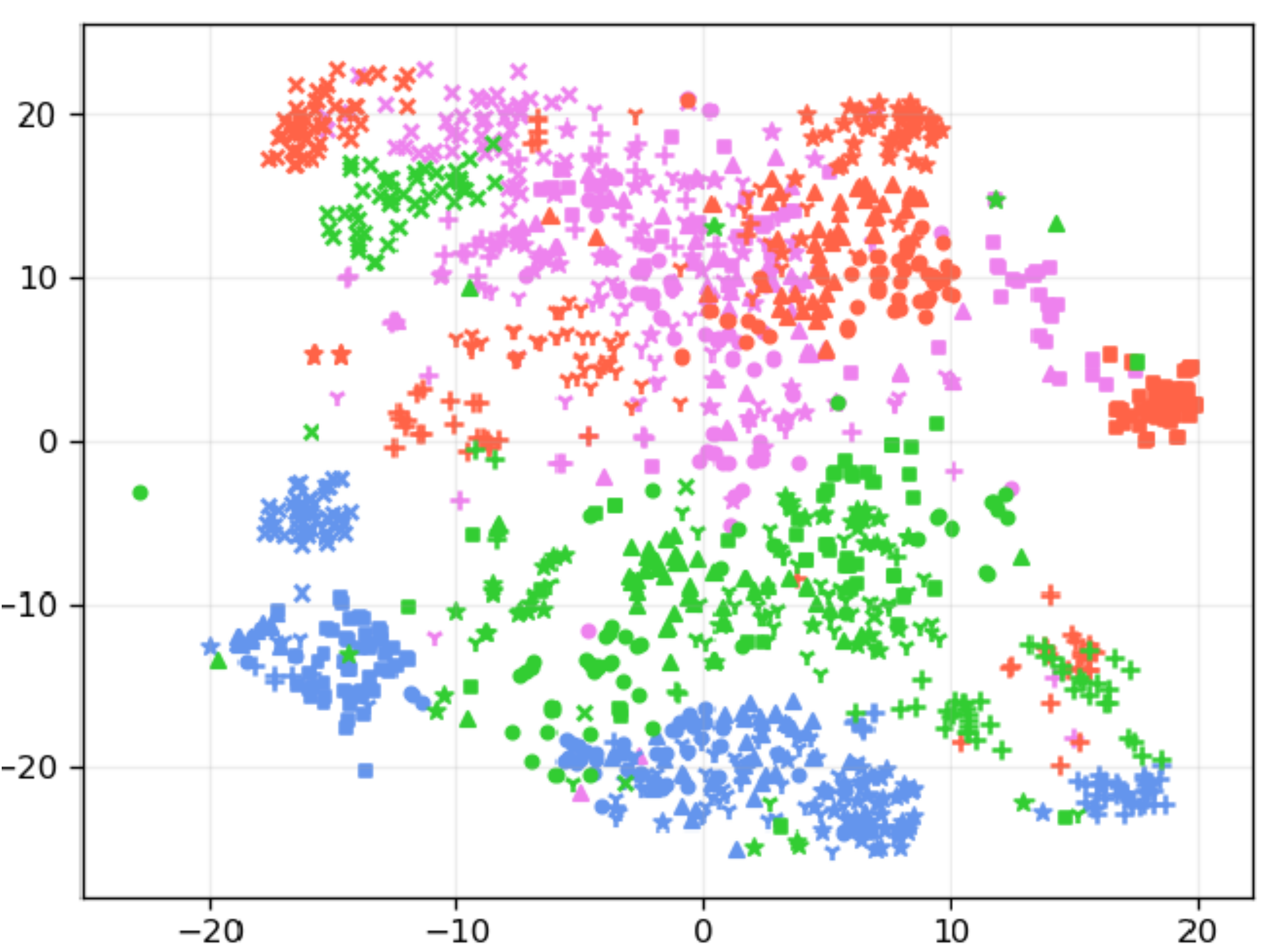}
\centering{(d)  Pre-trained AlexNet}
\end{minipage}
\begin{minipage}{0.33\textwidth}
\includegraphics[width=1\textwidth]{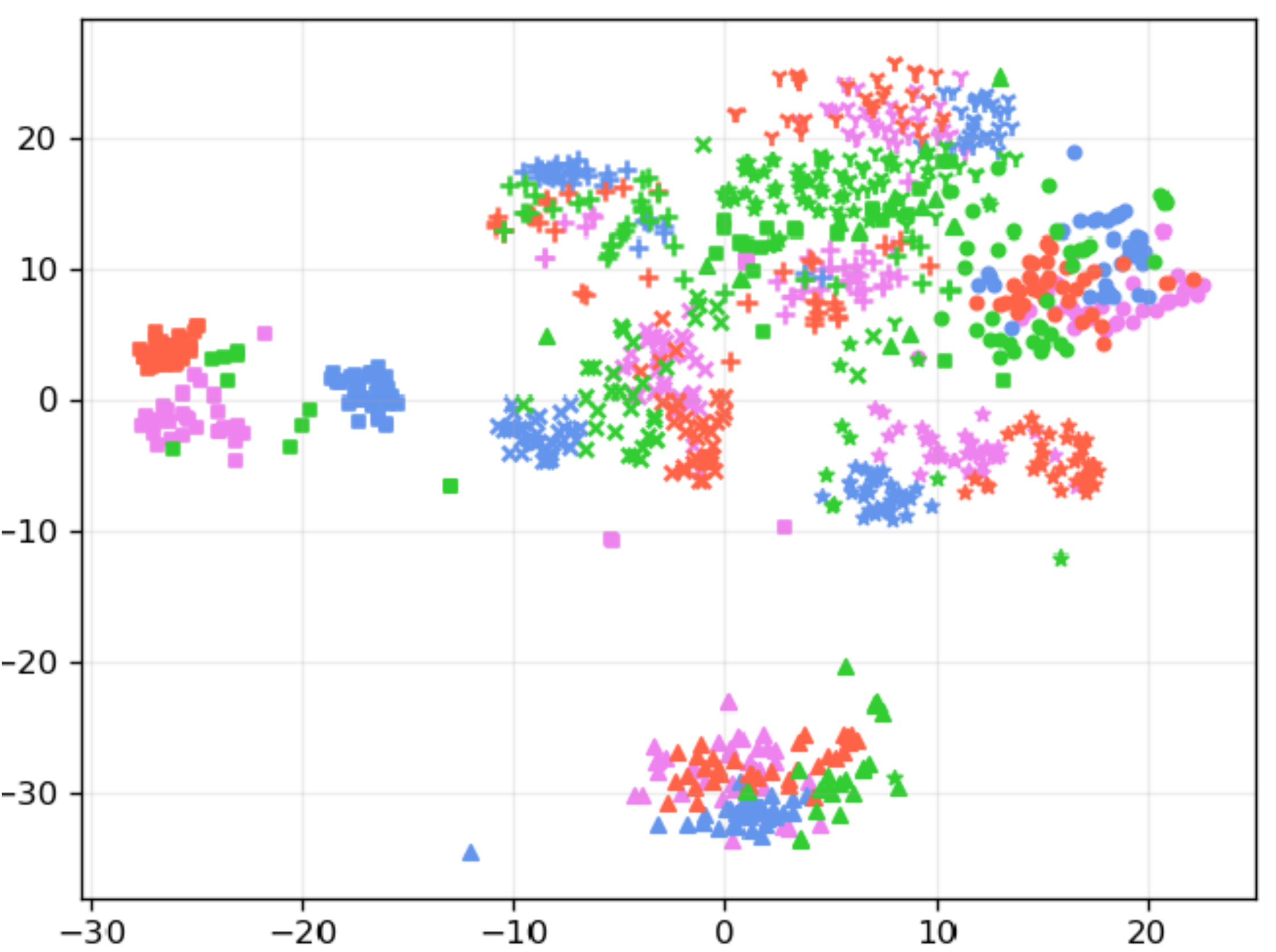}
\centering{(e) VIB}
\end{minipage}
\begin{minipage}{0.33\textwidth}
\includegraphics[width=1\textwidth]{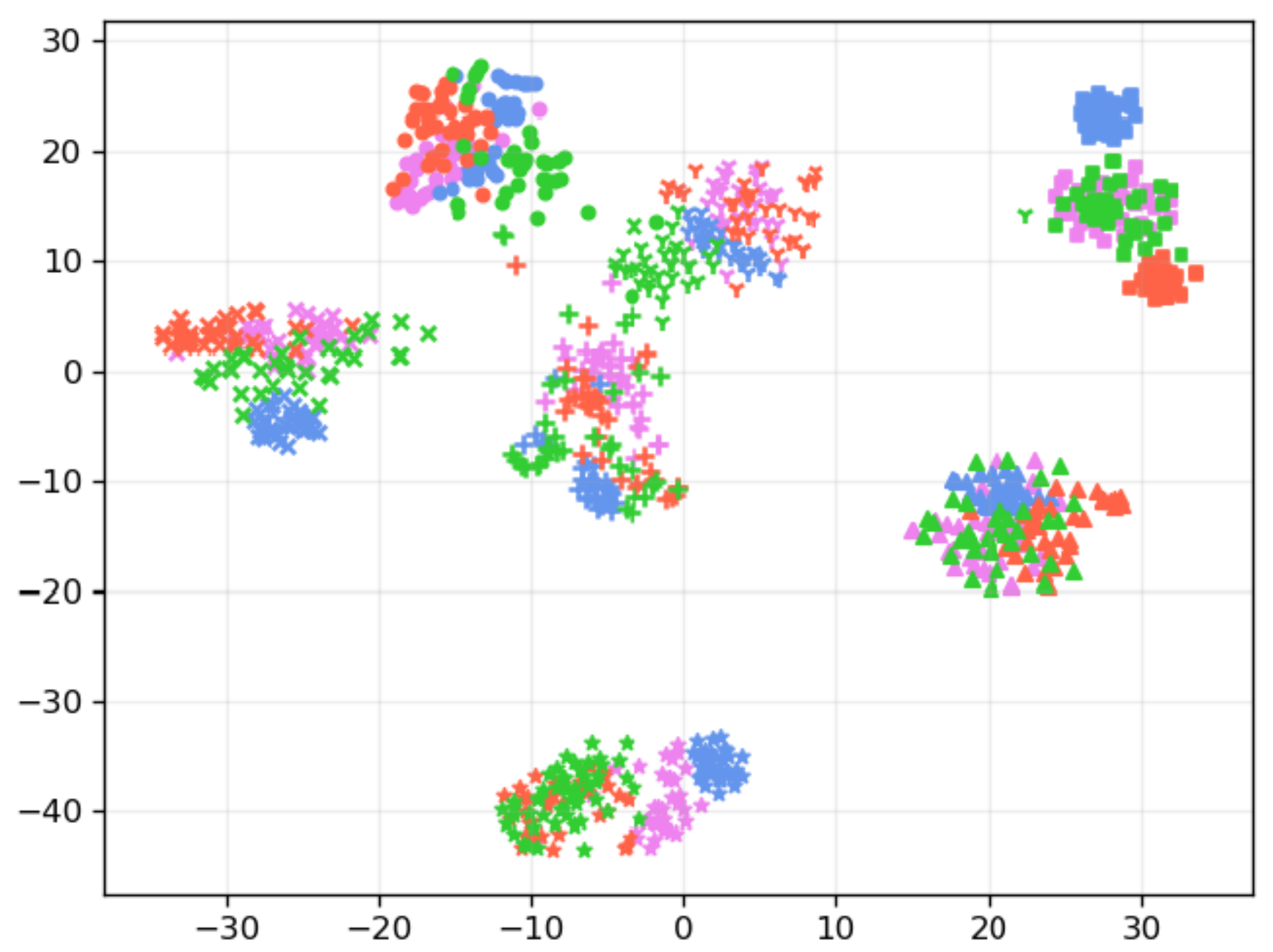}
\centering{(f) MetaVIB}
\end{minipage}
}
\caption{\textbf{Analyzing domain-invariance.} Visualization of feature representations from pre-trained AlexNet, VIB, and MetaVIB on PACS. The top row shows different features for the \textit{horse} category from four different domains, where the violet shapes denotes the unseen domain \textit{cartoon}. Bottom row 
shows the distributions of feature representations from all seven PACS classes for four domains, where the unseen domain (green) is \textit{art painting}.
MetaVIB reduces the domain gap to achieve domain-invariant yet discriminative representations, which enables accurate predictions. MetaVIB fills the gap between domains (c), while maximally separating samples of different classes (f).
}
\label{fig:tsne}
\end{center}

\end{figure*}

\subsubsection{Influence of information bottleneck size $\beta$} The bottleneck size $\beta$ controls the amount of information flow that goes through the bottleneck of the networks. To measure its influence on the performance, we plot the information plane dynamics of different network layers with varying $\beta$ in Fig.~\ref{fig:beta_ip}. We observe that MetaVIB with $\beta = 0.01$ achieves the highest $I(Z;Y)$ while at the same time $I(Z;X)$ is minimal. 
We also report the influence of $\beta$ in Table~\ref{tab:beta}. MetaVIB achieves the best performance when $\beta = 0.01$, which is consistent with the information dynamic in Fig.~\ref{fig:beta_ip}. We observe in Fig.~\ref{fig:beta_ip} (c) that with $\beta = 0.01$, the $I(X;T)$ is lowest and $I(T;Y)$ is the highest, compared to those with other values of $\beta$. A larger $I(Z;Y)$ indicates that we can make more accurate predictions $Y$ from $Z$, while a smaller $I(Z;X)$ indicates $Z$ contains the minimal information from $X$ that is required for prediction, suggesting a domain-invariant representation $Z$. This explains why $\beta = 0.01$ produces the best prediction results compared to other values of $\beta$. In our experiments, the optimal value of $\beta$ is obtained by using a validation set for each dataset and we found $\beta = 0.01$ produces the best performance on all datasets.

\subsubsection{Analyzing domain-invariance} We visualize the features learned by the pre-trained AlexNet, VIB and MetaVIB in Fig.~\ref{fig:tsne}. For better illustration, we use t-SNE~\cite{maaten2008visualizing} to reduce the feature dimension into a two-dimensional subspace. 
We observe that the features of the same category learned by pre-trained Alexnet (Fig.~\ref{fig:tsne} (a)) show large discrepancy among the four domains. The regular VIB reduces this discrepancy to some extent, but still suffers from considerable gaps between the unseen domain (violet shapes) (Fig.~\ref{fig:tsne} (b)). MetaVIB largely reduces the discrepancy of different domains including the unseen domains as shown in Fig.~\ref{fig:tsne} (c). In Fig.~\ref{fig:tsne} (d), we observe again that the gaps of features among 4 domains by the pre-trained AlexNet are larger than those between the 7 classes in each domain. Fig.~\ref{fig:tsne} (e) shows that the VIB reduces the domain gaps to certain extent. From Fig.~\ref{fig:tsne} (f), we observe MetaVIB reduces domain gaps considerably while at the same time scatters the samples of 7 classes in each domain. Overall, the proposed MetaVIB principle demonstrates effectiveness in learning domain-invariant representations to tackle domain shift.

\subsubsection{Success and failure cases} We show some success and failure cases in Fig.~\ref{fig:case}. MetaVIB successfully predicts the labels for ambiguous images. The dog in the second image in Fig.~\ref{fig:case} (a) wears human clothes, showing strong characteristics of a person. Yet, MetaVIB correctly predicts it with a high confidence probability of $0.732$. The sketch of the horse looks like a dog in the fourth image, but MetaVIB predicts it correctly with a high probability of $0.636$. In the failure cases (b), MetaVIB fails to make the correct prediction, but provides reasonable probabilities for both a person and a dog, which shows the effectiveness in handling uncertainty. It is hard to distinguish which object needs to be predicted in these images, as shown in the first image in Fig.~\ref{fig:case} (b).

\begin{figure*}[t]
\begin{center}
\subfigure{
\begin{minipage}{0.48\textwidth}
\includegraphics[width=1\textwidth]{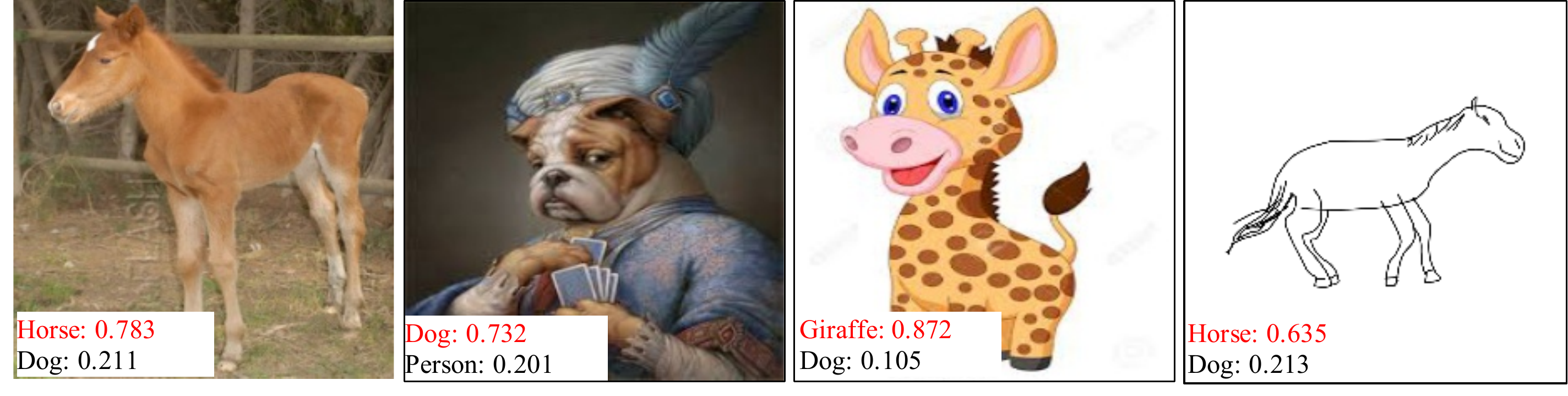}
\centering{(a) Success cases}
\end{minipage}
\begin{minipage}{0.48\textwidth}
\includegraphics[width=1\textwidth]{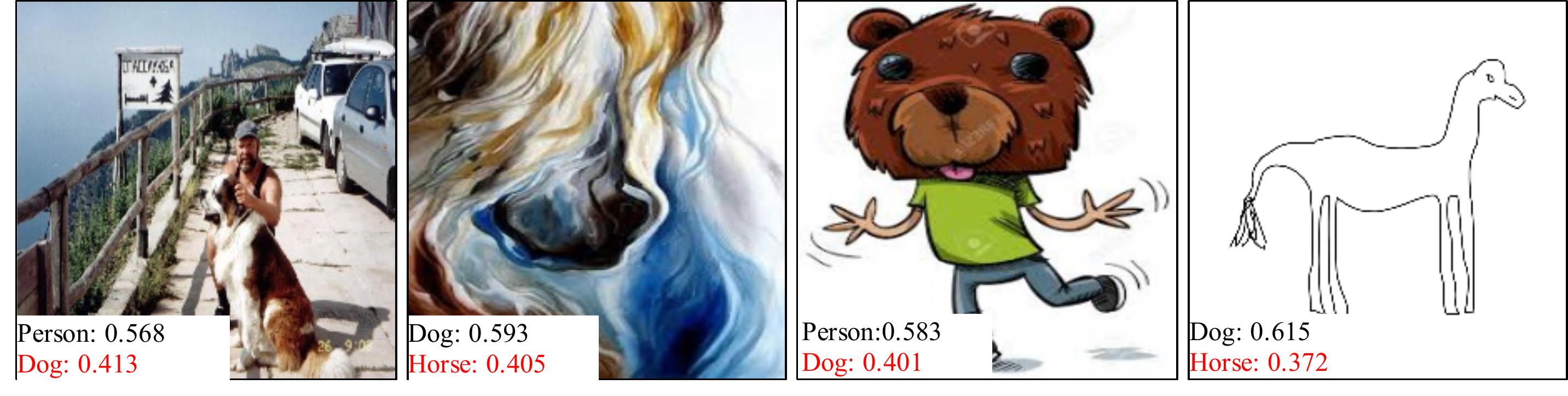}
\centering{(b) Failure cases}
\end{minipage}
}

\caption{\textbf{Success and failure cases} of MetaVIB. The numbers associated with each image are the top two prediction probabilities of MetaVIB, with ground truth labels in red. MetaVIB successfully distinguishes ambiguous cases in (a). For more challenging cases in (b), MetaVIB provides a high probability for the true label, but fails to make the correct prediction.}
\label{fig:case}
\end{center}

\end{figure*}

\subsection{State-of-the-Art Comparison}

\begin{table*}[t]
\small
\centering
\caption{\textbf{State-of-the-Art comparison} on VLCS, in classification accuracy (\%).}
\label{tab:vlcs}

 \makebox[\linewidth]{\begin{tabular}{l l l l l l}
\toprule
 & \textbf{VOC2007} & \textbf{LabelMe} & \textbf{Caltech-101} & \textbf{SUN09} & \textbf{Mean} \\
\midrule 
D`Innocente \& Caputo~\cite{d2018domain} & 66.06& 57.45 & 94.95  & 65.87 & 71.08 \\
Li et al.~\cite{li2017deeper}  & 69.99  & 63.49& 93.63 & 61.32 & 72.11  \\
Li et al.~\cite{li2018learning} & 67.70 & 62.60 & 94.40& 64.40 & 72.28  \\
Li et al.~\cite{Li_2019_ICCV}&  67.10  & 64.30 & 94.10 & 65.90  & 72.90 \\ 
Carlucci et al.~\cite{carlucci2019domain}
& \textbf{70.62}  & 60.90 & 96.93  & 64.30 & 73.19\\ 
Dou et al.\cite{dou2019domain}& 69.14 & \textbf{64.90} & 94.78 & 67.64 & 74.11 \\
\textbf{MetaVIB} &70.28{\scriptsize$\pm$0.21}  &62.66{\scriptsize$\pm$0.35}    &\textbf{97.37}{\scriptsize$\pm$0.23}  &\textbf{67.85}{\scriptsize$\pm$0.17} & \textbf{74.54} \\
\bottomrule
\end{tabular}}
\end{table*}

\begin{table*}[t]
\small
\centering
\caption{\textbf{State-of-the-Art comparison} on PACS, in classification accuracy (\%).}
\label{tab:pacs}

\begin{tabular}{l l l l l l}
\toprule
& \textbf{Photo} & \textbf{Art painting} & \textbf{Cartoon}  & \textbf{Sketch} & \textbf{Mean}\\
\midrule
Ghifary et al.~\cite{ghifary2015domain}& 91.12 & 60.27 & 58.65 & 47.68& 64.48  \\
Bousmalis et al.~\cite{bousmalis2016domain} & 83.25 & 61.13 & 66.54 & 58.58& 67.37\\
Li et al.~\cite{li2017deeper} & 89.50 & 62.86 & 66.97     & 57.51& 69.21 \\
Shankar et al.~\cite{shankar2018generalizing} & 89.48  & 64.84 & 67.69    & 57.52& 69.63 \\
Li et al.~\cite{li2018learning} & 88.00  & 66.23 & 66.88 & 58.96 & 70.01 \\
Nichol et al.~\cite{nichol2018first}& 88.78  & 64.35 & 70.09  & 59.91 & 70.78\\
Li et al.~\cite{Li_2019_ICCV}& 86.10 & 64.70  & 72.30  & 65.00  & 72.00 \\
Li et al.~\cite{li2019feature} & 89.94 & 64.89   & 71.72    & 61.85  &72.10\\
Balaji et al.~\cite{balaji2018metareg} & 91.70 & 69.82 & 70.35  & 59.26 & 72.62 \\
Carlucci et al.~\cite{carlucci2019domain} & 89.00 &  67.63  & 71.71  & 65.18  & 73.38 \\ 
Dou et al.\cite{dou2019domain}& 90.68 & 70.35 & 72.46 & \textbf{67.33} & 75.21 \\

\textbf{MetaVIB}&\textbf{91.93}{\scriptsize$\pm$0.23} &\textbf{71.94}{\scriptsize$\pm$0.34}  & \textbf{73.17}{\scriptsize$\pm$0.21}  &65.94{\scriptsize$\pm$0.24} & \textbf{75.74}   \\
\bottomrule
\end{tabular}

\end{table*}

\begin{table*}[t]
\scriptsize
\centering
\caption{\textbf{State-of-the-Art comparison} on Rotated MNIST, in averaged classification accuracy (\%) of different methods over 10 runs. MetaVIB consistently achieves the best performance on different domains with different rotation angles.}

 \makebox[\linewidth]{\begin{tabular}{lccccccc}
\toprule
 &$\mathbf{M_{0^{\circ}}}$ & $\mathbf{M_{15^{\circ}}}$ & $\mathbf{M_{30^{\circ}}}$ & $\mathbf{M_{45^{\circ}}}$ & $\mathbf{M_{60^{\circ}}}$ & $\mathbf{M_{75^{\circ}}}$ & \textbf{Mean} 
\\
\midrule
Shankar et al.\cite{shankar2018generalizing} & 86.03{\scriptsize$\pm$0.69} & 98.92{\scriptsize$\pm$0.53} &98.60{\scriptsize$\pm$0.51} &98.38{\scriptsize$\pm$0.29}& 98.68{\scriptsize$\pm$0.28} &88.94{\scriptsize$\pm$0.47}&94.93
\\
Balaji et al.\cite{balaji2018metareg} & 85.70{\scriptsize$\pm$0.31} &98.87{\scriptsize$\pm$0.41} &98.32{\scriptsize$\pm$0.44} &98.58{\scriptsize$\pm$0.28} &98.93{\scriptsize$\pm$0.32}&89.44{\scriptsize$\pm$0.37} & 94.97
\\
Li et al.\cite{li2018domain} &86.42{\scriptsize$\pm$0.24} &98.61{\scriptsize$\pm$0.27} &99.19{\scriptsize$\pm$0.19} &98.22{\scriptsize$\pm$0.24} &99.48{\scriptsize$\pm$0.19} &88.92{\scriptsize$\pm$0.43} &95.15
\\
Nichol et al.\cite{nichol2018first} &87.78{\scriptsize$\pm$0.30} &99.44{\scriptsize$\pm$0.22} &98.42{\scriptsize$\pm$0.24} &98.80{\scriptsize$\pm$0.20} &99.03{\scriptsize$\pm$0.28} &87.42{\scriptsize$\pm$0.33} &95.15
\\
Li et al.\cite{li2019feature} &89.23{\scriptsize$\pm$0.25} &99.68{\scriptsize$\pm$0.24} &99.20{\scriptsize$\pm$0.20} &99.24{\scriptsize$\pm$0.18} &99.53{\scriptsize$\pm$0.23} &91.44{\scriptsize$\pm$0.34} &96.39
\\
 \textbf{MetaVIB} & \textbf{91.28}{\scriptsize$\pm$0.21}&\textbf{99.90}{\scriptsize$\pm$0.02} &\textbf{99.29}{\scriptsize$\pm$0.11} &\textbf{99.78}{\scriptsize$\pm$0.10} &\textbf{99.57}{\scriptsize$\pm$0.13}&\textbf{92.75}{\scriptsize$\pm$0.31} & \textbf{97.08}\\
\bottomrule
\end{tabular}}
\label{tab:rMNIST}
\end{table*}
We compare with regular and meta-learning methods for domain generalization. The results on the three datasets are reported in Tables~\ref{tab:vlcs}-\ref{tab:rMNIST}.
On the \textbf{VLCS} dataset~\cite{vlcs}, our MetaVIB achieves high recognition accuracy, surpassing the second best method, i.e., MASF~\cite{dou2019domain}, by a margin of $0.43\%$.\ Note that on all domains, our MetaVIB consistently outperforms MLDG~\cite{li2018learning}, which is a gradient-based meta-learning algorithm.
On the \textbf{PACS} dataset~\cite{li2017deeper}, our MetaVIB again achieves the best overall performance.\ It outperforms most of the previous methods, showing clear performance advantages over JiGen~\cite{carlucci2019domain}. Again, our MetaVIB performs better than other meta-learning based methods, e.g., MetaReg~\cite{balaji2018metareg}, Reptile~\cite{nichol2018first}, MLDG~\cite{li2018learning}, Feature-Critic~\cite{li2019feature}, and MASF~\cite{dou2019domain}. It is worth highlighting that our MetaVIB exceeds those meta-learning methods on the ``Cartoon'' domain by phenomenal margins.
On the \textbf{Rotated MNIST} dataset~\cite{shankar2018generalizing}, the proposed MetaVIB achieves consistently high performance on the test domains, exceeding the alternative methods.\ It is worthwhile to mention that our MetaVIB outperforms the meta-learning algorithms MetaReg~\cite{balaji2018metareg}, and Reptile~\cite{nichol2018first}. showing its effectiveness as a meta-learning method for domain generalization.
To conclude, on all datasets, our MetaVIB accomplishes better performance than previous methods based on both regular learning and meta-learning.\ The best results on all benchmarks validate the effectiveness of our method for domain generalization.

\section{Conclusion}
In this work, we propose a new probabilistic model for domain generalization under the meta-learning framework. To address prediction uncertainty, we model the parameters of the classifiers shared across domains by a probabilistic distribution, which is inferred from the source domain and directly used for the target domains. To reduce domain shift, our method learns domain-invariant representations by a new Meta Variational Information Bottleneck principle, derived from a variational bound of mutual information. MetaVIB integrates the strengths of meta-learning, variational inference and probabilistic modeling for domain generalization. 
Our MetaVIB has been evaluated by extensive experiments on three benchmark datasets for cross-domain visual recognition. Ablation studies validate the benefits of our contributions. MetaVIB consistently achieves high performance and advances the state of the art on all three benchmarks. 
%
%
%
\bibliographystyle{splncs04}
\bibliography{MetaVIB}

\begin{thebibliography}{10}
\providecommand{\url}[1]{\texttt{#1}}
\providecommand{\urlprefix}{URL }
\providecommand{\doi}[1]{https://doi.org/#1}

\bibitem{alemi2016deep}
Alemi, A.A., Fischer, I., Dillon, J.V., Murphy, K.: Deep variational
  information bottleneck. In: International Conference on Learning
  Representations (2017)

\bibitem{amjad2019learning}
Amjad, R.A., Geiger, B.C.: Learning representations for neural network-based
  classification using the information bottleneck principle. IEEE Transactions
  on Pattern Analysis and Machine Intelligence  (2019)

\bibitem{andrychowicz2016learning}
Andrychowicz, M., Denil, M., Gomez, S., Hoffman, M.W., Pfau, D., Schaul, T.,
  Shillingford, B., de~Freitas, N.: Learning to learn by gradient descent by
  gradient descent. In: Advances in Neural Information Processing Systems
  (2016)

\bibitem{balaji2018metareg}
Balaji, Y., Sankaranarayanan, S., Chellappa, R.: Metareg: Towards domain
  generalization using meta-regularization. In: Advances in Neural Information
  Processing Systems. pp. 998--1008 (2018)

\bibitem{bertinetto2016learning}
Bertinetto, L., Henriques, J.F., Valmadre, J., Torr, P.H.S., Vedaldi, A.:
  Learning feed-forward one-shot learners. In: Advances in Neural Information
  Processing Systems (2016)

\bibitem{blanchard2011generalizing}
Blanchard, G., Lee, G., Scott, C.: Generalizing from several related
  classification tasks to a new unlabeled sample. In: Advances in neural
  information processing systems. pp. 2178--2186 (2011)

\bibitem{bousmalis2016domain}
Bousmalis, K., Trigeorgis, G., Silberman, N., Krishnan, D., Erhan, D.: Domain
  separation networks. In: Advances in Neural Information Processing Systems.
  pp. 343--351 (2016)

\bibitem{carlucci2019domain}
Carlucci, F.M., D'Innocente, A., Bucci, S., Caputo, B., Tommasi, T.: Domain
  generalization by solving jigsaw puzzles. In: IEEE Conference on Computer
  Vision and Pattern Recognition (2019)

\bibitem{choi2010exploiting}
Choi, M.J., Lim, J.J., Torralba, A., Willsky, A.S.: Exploiting hierarchical
  context on a large database of object categories. In: 2010 IEEE Computer
  Society Conference on Computer Vision and Pattern Recognition. pp. 129--136.
  IEEE (2010)

\bibitem{dou2019domain}
Dou, Q., de~Castro, D.C., Kamnitsas, K., Glocker, B.: Domain generalization via
  model-agnostic learning of semantic features. In: Advances in Neural
  Information Processing Systems. pp. 6447--6458 (2019)

\bibitem{d2018domain}
D’Innocente, A., Caputo, B.: Domain generalization with domain-specific
  aggregation modules. In: German Conference on Pattern Recognition. pp.
  187--198. Springer (2018)

\bibitem{erfani2016robust}
Erfani, S., Baktashmotlagh, M., Moshtaghi, M., Nguyen, V., Leckie, C., Bailey,
  J., Kotagiri, R.: Robust domain generalisation by enforcing distribution
  invariance. In: Proceedings of the International Joint Conference on
  Artificial Intelligence. pp. 1455--1461 (2016)

\bibitem{everingham2010pascal}
Everingham, M., Van~Gool, L., Williams, C.K., Winn, J., Zisserman, A.: The
  pascal visual object classes (voc) challenge. International journal of
  computer vision  \textbf{88}(2),  303--338 (2010)

\bibitem{finn2017model}
Finn, C., Abbeel, P., Levine, S.: Model-agnostic meta-learning for fast
  adaptation of deep networks. In: Proceedings of International Conference on
  Machine Learning. pp. 1126--1135 (2017)

\bibitem{finn2018}
Finn, C., Levine, S.: Meta-learning and universality: Deep representations and
  gradient descent can approximate any learning algorithm. In: International
  Conference on Learning Representations (2018)

\bibitem{finn2018probabilistic}
Finn, C., Xu, K., Levine, S.: Probabilistic model-agnostic meta-learning. In:
  Advances in Neural Information Processing Systems. pp. 9516--9527 (2018)

\bibitem{ghifary2015domain}
Ghifary, M., Bastiaan~Kleijn, W., Zhang, M., Balduzzi, D.: Domain
  generalization for object recognition with multi-task autoencoders. In: The
  IEEE International Conference on Computer Vision. pp. 2551--2559 (2015)

\bibitem{gordon2018meta}
Gordon, J., Bronskill, J., Bauer, M., Nowozin, S., Turner, R.E.: Meta-learning
  probabilistic inference for prediction. arXiv preprint arXiv:1805.09921
  (2018)

\bibitem{griffin2007caltech}
Griffin, G., Holub, A., Perona, P.: Caltech-256 object category dataset  (2007)

\bibitem{kingma2014adam}
Kingma, D.P., Ba, J.: Adam: A method for stochastic optimization. arXiv
  preprint arXiv:1412.6980  (2014)

\bibitem{kingma2013auto}
Kingma, D.P., Welling, M.: Auto-encoding variational bayes. In: International
  Conference on Learning Representations (2014)

\bibitem{kolchinsky2018caveats}
Kolchinsky, A., Tracey, B.D., Van~Kuyk, S.: Caveats for information bottleneck
  in deterministic scenarios. arXiv preprint arXiv:1808.07593  (2018)

\bibitem{krizhevsky2012imagenet}
Krizhevsky, A., Sutskever, I., Hinton, G.E.: Imagenet classification with deep
  convolutional neural networks. In: Advances in neural information processing
  systems. pp. 1097--1105 (2012)

\bibitem{lecun1998gradient}
LeCun, Y., Bottou, L., Bengio, Y., Haffner, P., et~al.: Gradient-based learning
  applied to document recognition. Proceedings of the IEEE  \textbf{86}(11),
  2278--2324 (1998)

\bibitem{li2017deeper}
Li, D., Yang, Y., Song, Y.Z., Hospedales, T.M.: Deeper, broader and artier
  domain generalization. In: Proceedings of the IEEE International Conference
  on Computer Vision. pp. 5542--5550 (2017)

\bibitem{li2018learning}
Li, D., Yang, Y., Song, Y.Z., Hospedales, T.M.: Learning to generalize:
  Meta-learning for domain generalization. In: Thirty-Second AAAI Conference on
  Artificial Intelligence (2018)

\bibitem{Li_2019_ICCV}
Li, D., Zhang, J., Yang, Y., Liu, C., Song, Y.Z., Hospedales, T.M.: Episodic
  training for domain generalization. In: IEEE International Conference on
  Computer Vision (2019)

\bibitem{li2018domain}
Li, H., Jialin~Pan, S., Wang, S., Kot, A.C.: Domain generalization with
  adversarial feature learning. In: Proceedings of the IEEE Conference on
  Computer Vision and Pattern Recognition. pp. 5400--5409 (2018)

\bibitem{li2018deep}
Li, Y., Tian, X., Gong, M., Liu, Y., Liu, T., Zhang, K., Tao, D.: Deep domain
  generalization via conditional invariant adversarial networks. In:
  Proceedings of the European Conference on Computer Vision. pp. 624--639
  (2018)

\bibitem{li2019feature}
Li, Y., Yang, Y., Zhou, W., Hospedales, T.M.: Feature-critic networks for
  heterogeneous domain generalization. In: Proceedings of International
  Conference on Machine Learning (2019)

\bibitem{maaten2008visualizing}
Maaten, L.v.d., Hinton, G.: Visualizing data using t-sne. Journal of machine
  learning research  \textbf{9}(Nov),  2579--2605 (2008)

\bibitem{muandet2013domain}
Muandet, K., Balduzzi, D., Sch{\"o}lkopf, B.: Domain generalization via
  invariant feature representation. In: Proceedings of Proceedings of
  International Conference on Machine Learning. pp. 10--18 (2013)

\bibitem{munkhdalai2017meta}
Munkhdalai, T., Yu, H.: Meta networks. In: Proceedings of International
  Conference on Machine Learning (2017)

\bibitem{nichol2018first}
Nichol, A., Achiam, J., Schulman, J.: On first-order meta-learning algorithms.
  arXiv preprint arXiv:1803.02999  (2018)

\bibitem{peng2018variational}
Peng, X.B., Kanazawa, A., Toyer, S., Abbeel, P., Levine, S.: Variational
  discriminator bottleneck: Improving imitation learning, inverse rl, and gans
  by constraining information flow. arXiv preprint arXiv:1810.00821  (2018)

\bibitem{ravi2017optimization}
Ravi, S., Larochelle, H.: Optimization as a model for few-shot learning. In:
  International Conference on Learning Representations (2017)

\bibitem{rezende2014stochastic}
Rezende, D.J., Mohamed, S., Wierstra, D.: Stochastic backpropagation and
  approximate inference in deep generative models. arXiv preprint
  arXiv:1401.4082  (2014)

\bibitem{russell2008labelme}
Russell, B.C., Torralba, A., Murphy, K.P., Freeman, W.T.: Labelme: a database
  and web-based tool for image annotation. International journal of computer
  vision  \textbf{77}(1-3),  157--173 (2008)

\bibitem{satorras2018few}
Satorras, V.G., Estrach, J.B.: Few-shot learning with graph neural networks.
  In: International Conference on Learning Representations (2018)

\bibitem{saxe2018information}
Saxe, A.M., Bansal, Y., Dapello, J., Advani, M., Kolchinsky, A., Tracey, B.D.,
  Cox, D.D.: On the information bottleneck theory of deep learning. In:
  International Conference on Learning Representations (2018)

\bibitem{Schmidhuber1992}
Schmidhuber, J.: Learning to control fast-weight memories: An alternative to
  dynamic recurrent networks. Neural Computation  \textbf{4}(1),  131--139
  (1992)

\bibitem{schmidhuber1997shifting}
Schmidhuber, J., Zhao, J., Wiering, M.: Shifting inductive bias with
  success-story algorithm, adaptive levin search, and incremental
  self-improvement. Machine Learning  \textbf{28}(1),  105--130 (1997)

\bibitem{shankar2018generalizing}
Shankar, S., Piratla, V., Chakrabarti, S., Chaudhuri, S., Jyothi, P., Sarawagi,
  S.: Generalizing across domains via cross-gradient training. arXiv preprint
  arXiv:1804.10745  (2018)

\bibitem{shwartz2017opening}
Shwartz-Ziv, R., Tishby, N.: Opening the black box of deep neural networks via
  information. arXiv preprint arXiv:1703.00810  (2017)

\bibitem{snell2017prototypical}
Snell, J., Swersky, K., Zemel, R.: Prototypical networks for few-shot learning.
  In: Advances in Neural Information Processing Systems. pp. 4077--4087 (2017)

\bibitem{sung2018learning}
Sung, F., Yang, Y., Zhang, L., Xiang, T., Torr, P.H., Hospedales, T.M.:
  Learning to compare: Relation network for few-shot learning. In: IEEE
  Conference on Computer Vision and Pattern Recognition. pp. 1199--1208 (2018)

\bibitem{thrun2012learning}
Thrun, S., Pratt, L.: Learning to learn. Springer Science \& Business Media
  (2012)

\bibitem{tishby2000ib}
Tishby, N., Pereira, F.C., Bialek, W.: The information bottleneck method. arXiv
  preprint physics/0004057  (2000)

\bibitem{Tishby2015dlib}
Tishby, N., Zaslavsky, N.: Deep learning and the information bottleneck
  principle. 2015 IEEE Information Theory Workshop (ITW)  (Apr 2015).
  \doi{10.1109/itw.2015.7133169}

\bibitem{vlcs}
Torralba, A., Efros, A.A., et~al.: Unbiased look at dataset bias. In: IEEE
  Conference on Computer Vision and Pattern Recognition (2011)

\bibitem{vilalta2002perspective}
Vilalta, R., Drissi, Y.: A perspective view and survey of meta-learning.
  Artificial Intelligence Review  \textbf{18}(2),  77--95 (2002)

\bibitem{Vinyals2016}
Vinyals, O., Blundell, C., Lillicrap, T., Kavukcuoglu, K., Wierstra, D.:
  Matching networks for one shot learning. In: Advances in Neural Information
  Processing Systems. pp. 3637--3645 (2016)

\bibitem{xie2017controllable}
Xie, Q., Dai, Z., Du, Y., Hovy, E., Neubig, G.: Controllable invariance through
  adversarial feature learning. In: Advances in Neural Information Processing
  Systems. pp. 585--596 (2017)

\bibitem{zhen2020learning}
Zhen, X., Sun, H., Du, Y., Xu, J., Yin, Y., Shao, L., Snoek, C.: Learning to
  learn kernels with variational random features. International Conference on
  Machine Learning  (2020)

\end{thebibliography}

\newpage
\appendix

\renewcommand\thefigure{\thesection.\arabic{figure}} 
\renewcommand\thetable{\thesection.\arabic{table}} 
\renewcommand\theequation{\thesection.\arabic{equation}} 

\setcounter{figure}{0}
\setcounter{table}{0}
\setcounter{equation}{0}

\section{Algorithms of MetaVIB for Training}
We describe the detailed algorithm for training MetaVIB as following Algorithm~\ref{algo}:

\begin{algorithm}[h]
\caption{Learning to Learn with Variational Information Bottleneck for Domain Generalization}
\label{algo}
\begin{algorithmic}[1]
\STATE \textbf{Input:} Training data $\mathcal{S}$ of $K$ source domains; learning rate $\lambda$; the number of iteration $N_{iter}$.
\STATE Initialize the parameters $\boldsymbol{\Theta}=\left\{\theta, \phi_1, \phi_2\right\}$ of the model including the feature extraction network $h_{\theta}(\cdot)$ and the inference networks $g_{\phi_1}(\cdot)$ and $g_{\phi_2}(\cdot)$.
\FOR{\textit{iter} in $N_{iter}$}
\STATE $D^t$ $\leftarrow$ RANDOMSAMPLE($\{1,\cdots,K\}$, $t$); 
\\
$D^s$ $\leftarrow$ $\{1,\cdots,K\}$ $\backslash$ $D^t$;
\STATE Sample $\{(\mathbf{x}_m^s,\mathbf{y}_m^s)\}^M_{m=1}\sim D^s$; $\{(\mathbf{x}^t_n, \mathbf{y}^t_n)\}^N_{n=1}\sim\mathcal{D}^{t}$;
\FOR{$c$ in $1:C$}
\STATE $\overline{\mathbf{h}}_{c}^{s} = \frac{1}{M_c} \sum\limits^{M_c}_{i=1}h_{\theta}(\mathbf{x}_{i, c}^{s})$; $\boldsymbol{{\mu}_c}^\psi, \boldsymbol{{\sigma}_c}^\psi = g_{\phi_1}(\overline{\mathbf{h}}_{c}^{s})$;
\\
$\psi_c \sim \mathcal{N}(\boldsymbol{{\mu}_c}^\psi, \text{diag}((\boldsymbol{{\sigma}_c}^\psi)^2)$;
\ENDFOR
\STATE $\psi = \left[\psi_1, \cdots, \psi_c, \cdots, \psi_C \right]$;
\FOR{$c$ in $ 1 : C$}
\STATE $\overline{\mathbf{h}}_{c}^s = \frac{1}{M_c} \sum\limits^{M_c}_{i=1}h_{\theta}(\mathbf{x}_{i, c}^{s})$; $\boldsymbol{\mu}_{c}^{s}, \boldsymbol{\sigma}_{c}^{s} = g_{\phi_2}(\overline{\mathbf{h}}_{c}^{s})$;\\ $\mathbf{z}_c \sim \mathcal{N}(\boldsymbol{\mu}_{c}^{s}, \text{diag}((\boldsymbol{\sigma}_{c}^{s})^2))$;
\STATE $\boldsymbol{\mu}^{t}_{j, c}, \boldsymbol{\sigma}^t_{j, c} = g_{\phi_2}(h_{\theta}(\mathbf{x}^t_{j, c}))$; 

$\mathbf{z}_{j, c} \sim \mathcal{N}(\boldsymbol{\mu}^t_{j, c},  \text{diag}((\boldsymbol{\sigma}^{t}_{j, c})^2))$;
\STATE \scriptsize{$\mathcal{L}_c={{{\sum\limits_{\scriptscriptstyle (\mathbf{x}^t_{j, c}, \mathbf{y}^t_c)}}}} \left[-\psi_{y} \cdot \mathbf{z}_{j, c} + \log (\sum\limits_{c=1}^{C}e^{\psi_c \cdot  \mathbf{z}_{j, c}})\right] + 
\beta D_{\mathrm{KL}}(q(\mathbf{z}_c|\overline{\mathbf{h}}_{c})|| p(\mathbf{z}_{j, c}|h_{\theta}(\mathbf{x}^t_{j, c})))$};
\ENDFOR
\STATE Update parameters: $\boldsymbol{\Theta} \leftarrow \boldsymbol{\Theta} -\lambda \sum\limits_{c=1}^C \nabla_{\Theta} \mathcal{L}_c$.
\ENDFOR
\end{algorithmic}
\end{algorithm}

\section{Learning Architecture}
\begin{figure*}[h]
\begin{center}
\includegraphics[width=.8\textwidth]{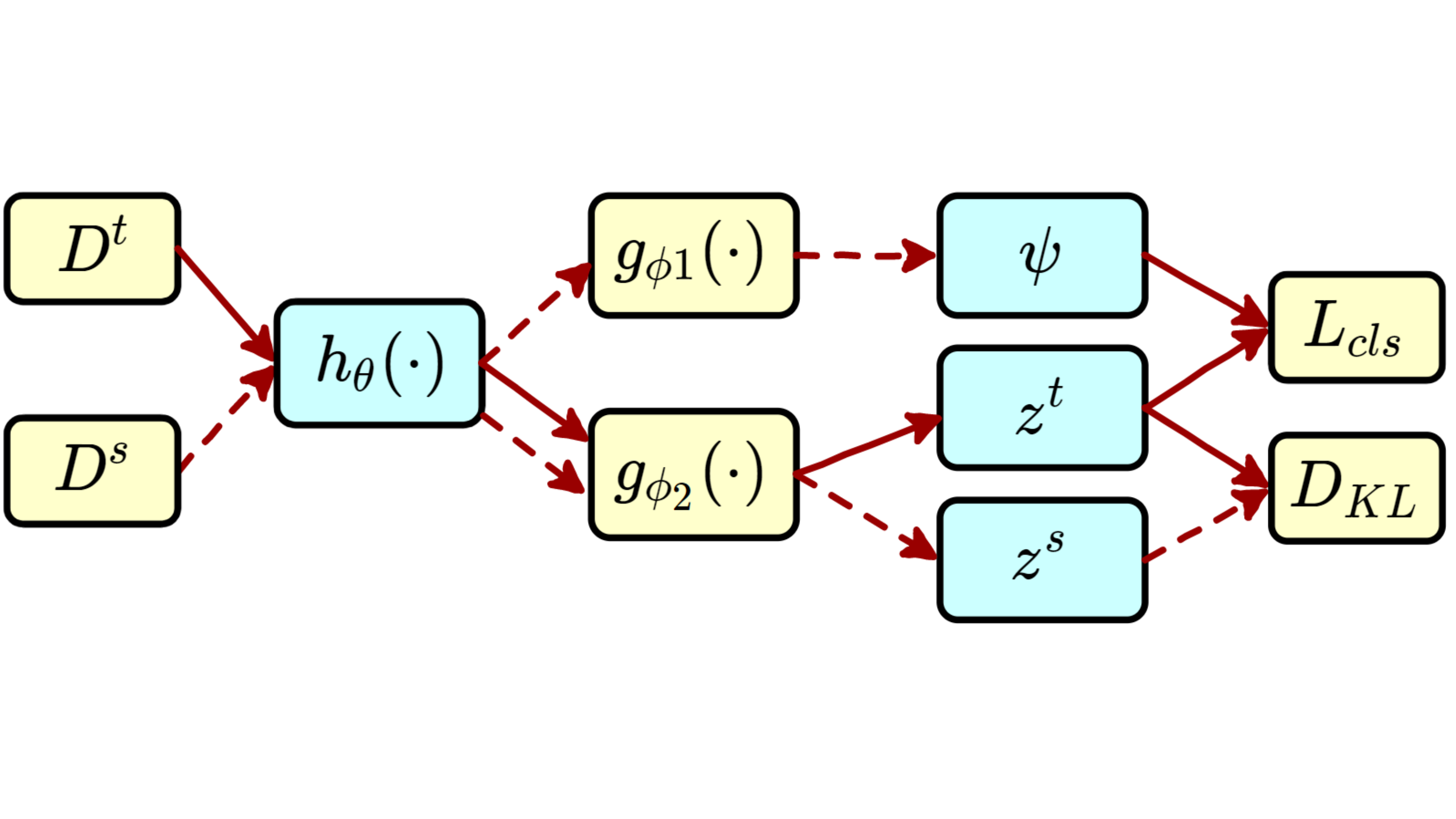}
\caption{\textbf{Architecture diagram}. $h_{\theta}(\cdot)$ is the feature extraction network; $g_{\phi_1}(\cdot)$ is the inference network to generate the distribution of classifier parameters $\psi$; $g_{\phi_2}(\cdot)$ is the inference network to generate the latent distribution of $z$; $L_{cls}$ is the cross-entropy loss. Solid (Dashed) line represents the direction of data flow in the meta-test domain $D^t$ (meta-train domain $D^s$).}
\label{fig:diagram}
\end{center}
\end{figure*}

To better clearly understand our proposed MetaVIB, we draw a concise architecture diagram in Fig.~\ref{fig:diagram}.

\section{Training Details}
During the training, we use the Adam~\cite{kingma2014adam} optimizer, and set the learning rate as $10^{-4}$.\ In each training batch, we randomly select three domains including two meta-train domains and one meta-test domain.\ In each domain, we choose $256$ samples, and the batch size is $256\times3$.\ The iteration number is set as $25,000$.\ The model with the highest validation accuracy is employed to evaluate the test set from the meta-test domain.

\section{Influence of information bottleneck size $\beta$}
We report Influence of information bottleneck size $\beta$ on the VLCS and Rotated MNIST in Tables~\ref{tab:beta_VLCS} and~\ref{tab:beta-rMNIST}. For the VLCS, MetaVIB obtains best results for $\beta = 0.01$, while for the Rotated MNIST, MetaVIB gets best results for $\beta = 0.001$.

\begin{table*}[h]
\setlength{\tabcolsep}{6pt}
\centering
\caption{\textbf{Influence of information bottleneck size $\beta$} on domain generalization for VLCS.} \label{tab:beta_VLCS}
\begin{tabular}{lccccc}
\toprule
 & \textbf{VOC2007} & \textbf{LabelMe} & \textbf{Caltech-101} & \textbf{SUN09} & \textbf{Mean} \\
\midrule
$\beta=1$& 67.15\scriptsize{$\pm$0.31} & 60.32\scriptsize{$\pm$0.37} & 94.83\scriptsize{$\pm$0.25} & 65.02\scriptsize{$\pm$0.23} & 71.83 \\
$\beta=0.1$ & 68.93\scriptsize{$\pm$0.24} & 61.31\scriptsize{$\pm$0.18} & 95.98\scriptsize{$\pm$0.21} & 67.05\scriptsize{$\pm$0.21}& 73.32 \\
$\beta=0.01$& 70.28\scriptsize{$\pm$0.34}  & 62.66\scriptsize{$\pm$0.24} & 97.37\scriptsize{$\pm$0.33} & 67.85\scriptsize{$\pm$0.27} & 74.54 \\
$\beta=0.001$& 68.47\scriptsize{$\pm$0.35} & 61.17\scriptsize{$\pm$0.27}  & 95.35\scriptsize{$\pm$0.23}  & 66.90\scriptsize{$\pm$0.25} & 72.97 \\
\bottomrule
\end{tabular}
\end{table*}

\begin{table*}[h]
\scriptsize
\centering
\caption{\textbf{Influence of information bottleneck size $\beta$} on domain generalization for Rotated MNIST.} 
 \makebox[\linewidth]{\begin{tabular}{lccccccc}
\toprule
 &$\mathbf{M_{0^{\circ}}}$ & $\mathbf{M_{15^{\circ}}}$ & $\mathbf{M_{30^{\circ}}}$ & $\mathbf{M_{45^{\circ}}}$ & $\mathbf{M_{60^{\circ}}}$ & $\mathbf{M_{75^{\circ}}}$ & \textbf{Mean} 
\\
\midrule

$\beta=1$ &89.13\scriptsize{$\pm$0.24} &98.01\scriptsize{$\pm$0.21} &97.38\scriptsize{$\pm$0.18} &97.32\scriptsize{$\pm$0.20} &98.13\scriptsize{$\pm$0.28} &88.72\scriptsize{$\pm$0.13} &94.78
\\
$\beta=0.1$ &90.35\scriptsize{$\pm$0.31} &98.17\scriptsize{$\pm$0.21} &98.82\scriptsize{$\pm$0.34} &98.18\scriptsize{$\pm$0.31} &98.73\scriptsize{$\pm$0.29}&89.94\scriptsize{$\pm$0.17} & 95.69
\\
$\beta=0.01$ & 91.05\scriptsize{$\pm$0.19} & 99.35\scriptsize{$\pm$0.03} &99.10\scriptsize{$\pm$0.31} &99.38\scriptsize{$\pm$0.18}& 99.27\scriptsize{$\pm$0.18} &91.94\scriptsize{$\pm$0.47}&96.68
\\
$\beta=0.001$ & 91.28\scriptsize{$\pm$0.21}&99.90\scriptsize{$\pm$0.02} &99.29\scriptsize{$\pm$0.11} &99.78\scriptsize{$\pm$0.10} &99.57\scriptsize{$\pm$0.13}&92.75\scriptsize{$\pm$0.31} & 97.08\\
\bottomrule
\end{tabular}}
\label{tab:beta-rMNIST}
\end{table*}

\section{Influence of the number of  Monte Carlo Influence of the number of Monte Carlo samples}
We use Monte Carlo sampling to draw samples from$p(\mathbf{z}|\mathbf{x})$ for $\mathbf{z}$. We report varying sample number $L_z$ on PACS in the Table~\ref{tab:MCS}. Our method achieves inferior results with $L_z=1$; performs consistently better with $L_z=5, 10$, converges at $L_z=10$ and becomes worse when $L_z=50, 100$. So in our experiments, we set $L_z=10$ and we averaged over $20$ runs on the test domain. The variance reflects the error caused by Monte Carlo sampling in each test experiment.

\begin{table*}[h]
\setlength{\tabcolsep}{6pt}
\centering
\caption{\textbf{Influence of the number of  Monte Carlo samples $L_z$} on domain generalization for PACS. MetaVIB obtains best results for $L_z=10$.} \label{tab:MCS}
\begin{tabular}{lccccc}
\toprule
& \textbf{Photo} & \textbf{Art painting} & \textbf{Cartoon} & \textbf{Sketch} & \textbf{Mean} \\
\midrule
$L_z=1$& 89.32{\scriptsize$\pm$0.41} & 69.17{\scriptsize$\pm$0.37} & 70.37{\scriptsize$\pm$0.27}  & 62.84{\scriptsize$\pm$0.45} & 72.93 \\
$L_z = 5$ & 90.11{\scriptsize$\pm$0.17} & 70.26{\scriptsize $\pm$0.38} & 71.93{\scriptsize$\pm$0.21} & 63.45{\scriptsize$\pm$0.46}& 73.94 \\
$L_z = 10$&91.93{\scriptsize$\pm$0.23} &71.94{\scriptsize$\pm$0.34}  & 73.17{\scriptsize$\pm$0.21}  &65.94{\scriptsize$\pm$0.24} & 75.74  \\
$L_z = 50$& 91.82{\scriptsize$\pm$0.25} & 71.74{\scriptsize$\pm$0.32}  & 73.37{\scriptsize$\pm$0.17}    & 66.01{\scriptsize$\pm$0.38} & 75.73 \\
$L_z = 100$& 91.71{\scriptsize$\pm$0.35} & 71.87{\scriptsize$\pm$0.37}  & 73.09{\scriptsize$\pm$0.27}    & 65.81{\scriptsize$\pm$0.48} & 75.62 \\
\bottomrule
\end{tabular}
\end{table*}

\section{Network Architectures}
\subsection{Feature Embedding Network}
The feature extraction network for \textbf{PACS}, \textbf{VLCS} is shown in Table \ref{pacs:h}, the feature extraction network for \textbf{Rotated MNIST} is shown in Table \ref{mnist:h}.
\subsection{Inference Network}
The architecture of the inference network $g_{\phi_1}(\cdot)$ for \textbf{PACS}, \textbf{VLCS} is in Table \ref{pacs:g1}, the architecture of the inference network $g_{\phi_1}(\cdot)$ for \textbf{Rotated MNIST} is in Table \ref{mnist:g1}.

The architecture of the inference network $g_{\phi_2}(\cdot)$ for \textbf{PACS}, \textbf{VLCS} is in Table \ref{pacs:g2}, the architecture of the inference network $g_{\phi_2}(\cdot)$ for \textbf{Rotated MNIST} is in Table \ref{mnist:g2}.

\begin{table}[h]
    \caption{The feature extraction network $h_{\theta}(\cdot)$ for \textbf{PACS}, \textbf{VLCS}}
	\centering
	\begin{tabular}{cl}
		\multicolumn{2}{l} {Feature Extraction Network : $ h_{\theta}(\cdot)$} \\
		\toprule
		\textbf{Output size} & \textbf{Layers} \\
        \midrule
		$227 \times 227 \times 3$ & Input image \\
        \hline
		$27 \times 27 \times 96$ & conv2d ($11 \times 11$, stride 4, SAME, RELU),  pool ($3 \times 3 $, stride 2, VALID) \\
		$13 \times 13 \times 256$ & conv2d ($5 \times 5$, stride 1, SAME, RELU),  pool ($3 \times 3 $, stride 2, VALID) \\
		$13 \times 13 \times 384$ & conv2d ($3 \times 3$, stride 1, SAME, RELU) \\
		$13 \times 13 \times 384$ & conv2d ($3 \times 3$, stride 1, SAME, RELU) \\
		$6 \times 6 \times 256$ & conv2d ($3 \times 3$, stride 1, SAME, RELU) ,  pool ($3 \times 3 $, stride 2, VALID)\\
		$4096$ & fully connected, RELU, dropout \\
		$4096$ & fully connected, RELU \\
        \bottomrule
	\end{tabular}
	\label{pacs:h}
\end{table}

\begin{table}[h]
    \caption{The feature extraction network $h_{\theta}(\cdot)$ for \textbf{Rotated MNIST}}
	\centering
	\begin{tabular}{cl}
		\multicolumn{2}{l} {Feature Extraction Network : $ h_{\theta}(\cdot)$} \\
		\toprule
		\textbf{Output size} & \textbf{Layers} \\
        \midrule
		$28 \times 28 \times 1$ & Input image \\
        \hline
		$14 \times 14 \times 32$ & conv2d ($3 \times 3$, stride 1, SAME, RELU),  pool ($3 \times 3 $, stride 2, VALID) \\
		$7 \times 7 \times 32$ & conv2d ($3 \times 3$, stride 1, SAME, RELU),  pool ($3 \times 3 $, stride 2, VALID) \\
		$256$ & fully connected, RELU \\
        \bottomrule
	\end{tabular}
	\label{mnist:h}
\end{table}

\begin{table}[h]
    \caption{Inference network $g_{\phi_1}(\cdot)$ used for \textbf{PACS}, \textbf{VLCS}.}
	\centering
	\begin{tabular}{cl}
	  \multicolumn{2}{l}{Inference Network: $g_{\phi_1}(\cdot)$} \\
      \toprule
    	\textbf{Output size} & \textbf{Layers} \\
        \midrule
		$k \times 4096$ & Input feature \\
        \hline
		$4096$ & instance pooling \\
		$1024$ & fully connected, ELU \\
		$1024$ & fully connected, ELU \\
		$1024$ & fully connected to $\mu_c^\psi$, $\log(\sigma^\psi_c)^2$ \\
	
      \bottomrule
	\end{tabular}
	\label{pacs:g1}
\end{table}

\begin{table}[h]
    \caption{Inference network $g_{\phi_1}(\cdot)$ used for \textbf{Rotated MNIST}.}
	\centering
	\begin{tabular}{cl}
	  \multicolumn{2}{l}{Inference Network: $g_{\phi_1}(\cdot)$} \\
      \toprule
    	\textbf{Output size} & \textbf{Layers} \\
        \midrule
		$k \times 256$ & Input feature \\
 \hline
		$256$ & instance pooling \\
		$256$ & fully connected, ELU \\
		$256$ & fully connected, ELU \\
		$256$ & fully connected to $\mu_c^\psi$, $\log(\sigma^\psi_c)^2$ \\
	
      \bottomrule
	\end{tabular}
	\label{mnist:g1}
\end{table}

\begin{table}[h]
    \caption{Inference network $g_{\phi_2}(\cdot)$ used for \textbf{PACS}, \textbf{VLCS}.}
	\centering
	\begin{tabular}{cl}
	  \multicolumn{2}{l}{Inference Network: $g_{\phi_2}(\cdot)$} \\
      \toprule
    	\textbf{Output size} & \textbf{Layers} \\
        \midrule
		$k \times 4096$ & Input feature \\
 \hline
		$4096$ & instance pooling \\
		$1024$ & fully connected, ELU \\
		$1024$ & fully connected, ELU \\
		$1024$ & fully connected to $\mu_c$, $\log(\sigma_c)^2$ \\
	
      \bottomrule
	\end{tabular}
	\label{pacs:g2}
\end{table}

\begin{table}[h]
    \caption{Inference network $g_{\phi_2}(\cdot)$ used for \textbf{Rotated MNIST}.}
	\centering
	\begin{tabular}{cl}
	  \multicolumn{2}{l}{Inference Network: $g_{\phi_2}(\cdot)$} \\
      \toprule
    	\textbf{Output size} & \textbf{Layers} \\
        \midrule
		$k \times 256$ & Input feature \\
 \hline
		$256$ & instance pooling \\
		$256$ & fully connected, ELU \\
		$256$ & fully connected, ELU \\
		$256$ & fully connected to $\mu_c$, $\log(\sigma_c)^2$ \\
	
      \bottomrule
	\end{tabular}
	\label{mnist:g2}
\end{table}

\section{Prediction Uncertainty Analysis}
Since the data follows distinct distribution between seen and unseen domains, uncertainty is inevitable during the prediction stage on the unseen domains, to which no data is accessible in the learning stage. To deal with the prediction uncertainty, we model parameters of classifiers shared across domains as probabilistic distributions that we infer from the data of the seen domains. The probabilistic modeling enables us to better handle the prediction uncertainty on previously unseen domains. 

\begin{figure*}[htb]
\begin{center}
\includegraphics[width=1.0\textwidth]{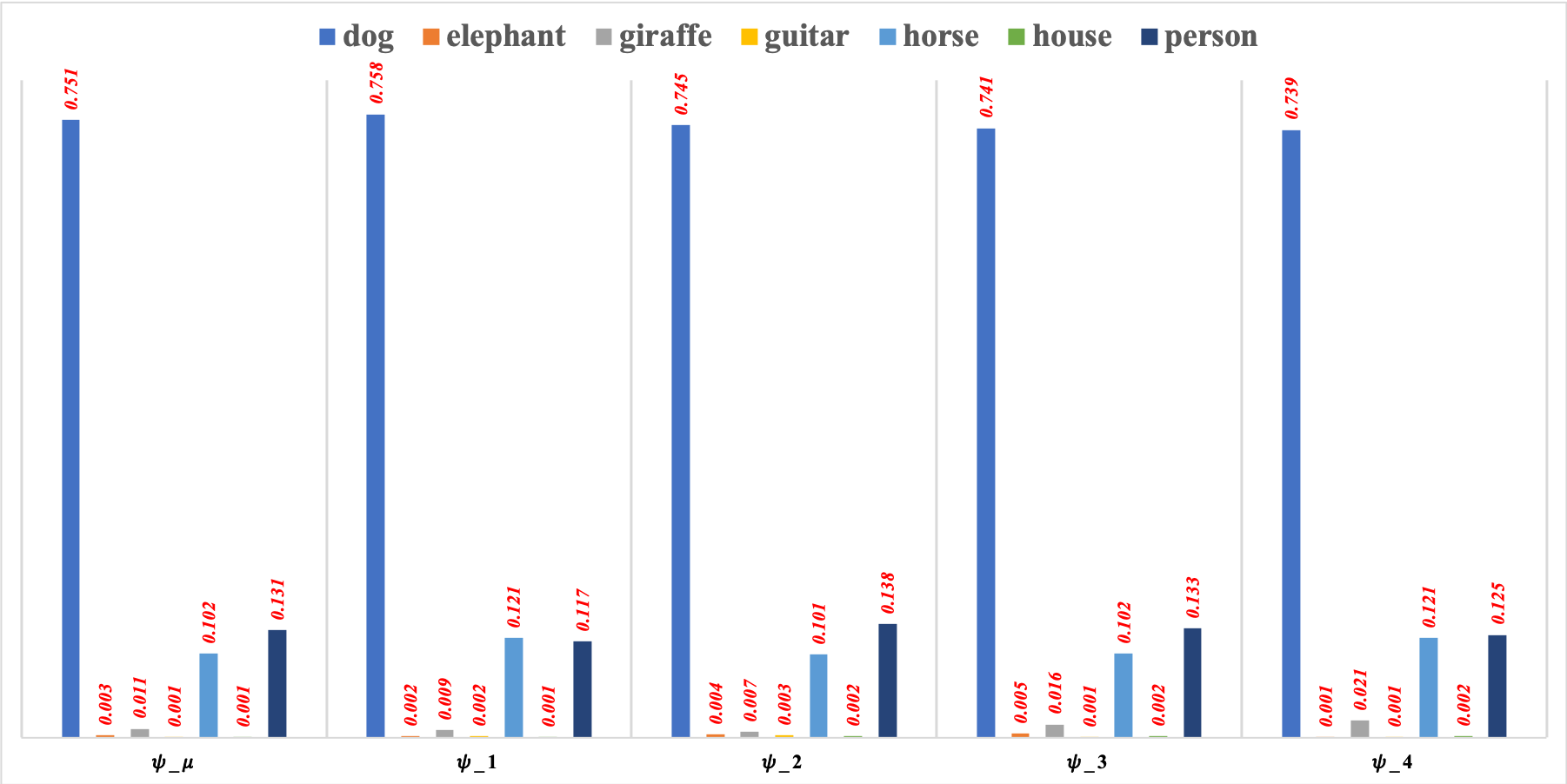}
\caption{The prediction probability of the different sampled classifier $\psi$ for each category (the first image of the success cases in Fig.~\ref{fig:case}).}
\label{fig:c1}
\end{center}
\end{figure*}

\begin{figure*}[htb]
\begin{center}
\includegraphics[width=1.0\textwidth]{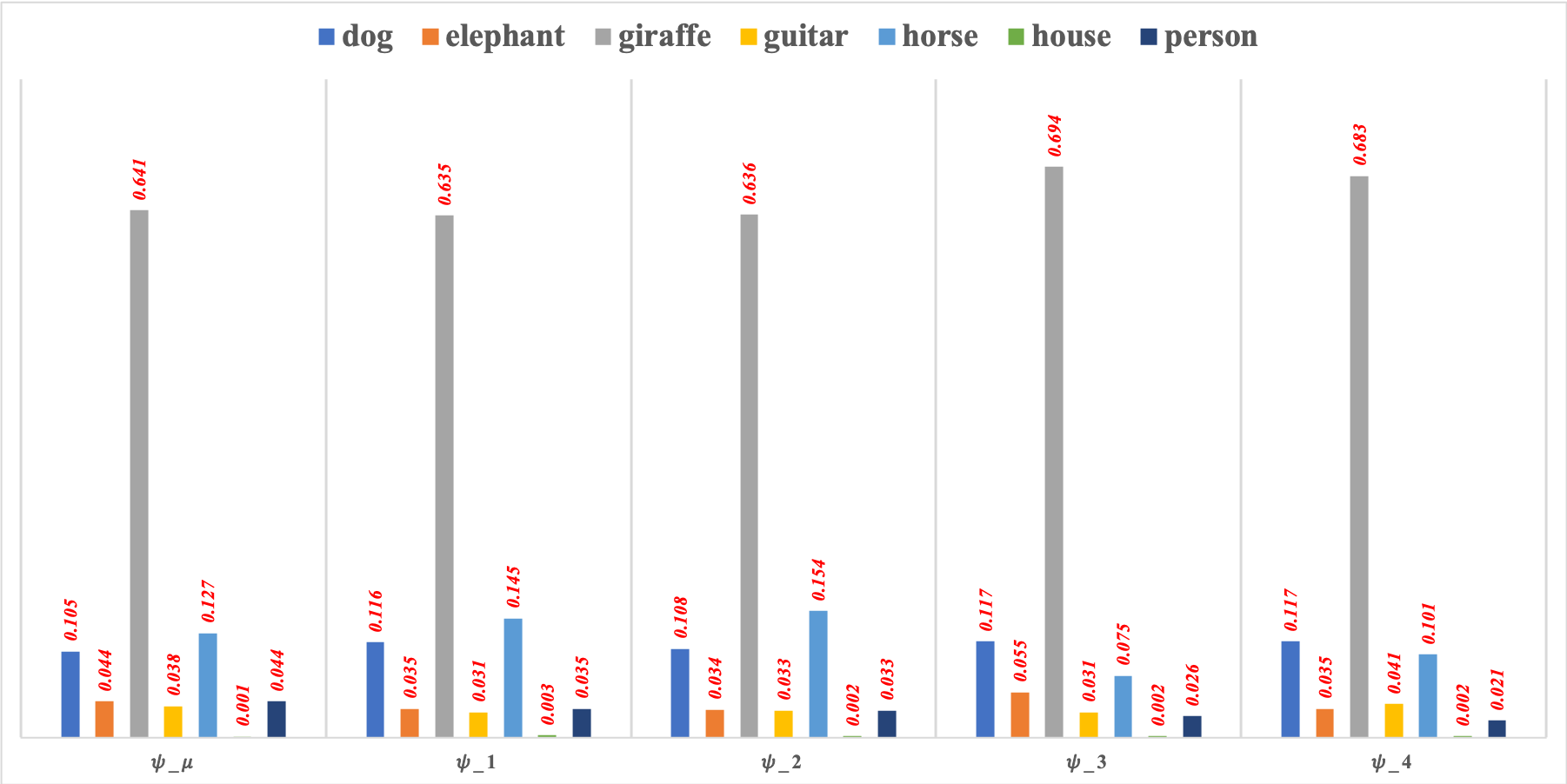}
\caption{The prediction probability of the different sampled classifier $\psi$ for each category (the Second image of the success cases in Fig.~\ref{fig:case}).}
\label{fig:c2}
\end{center}
\end{figure*}

\begin{figure*}[htb]
\begin{center}
\includegraphics[width=1.0\textwidth]{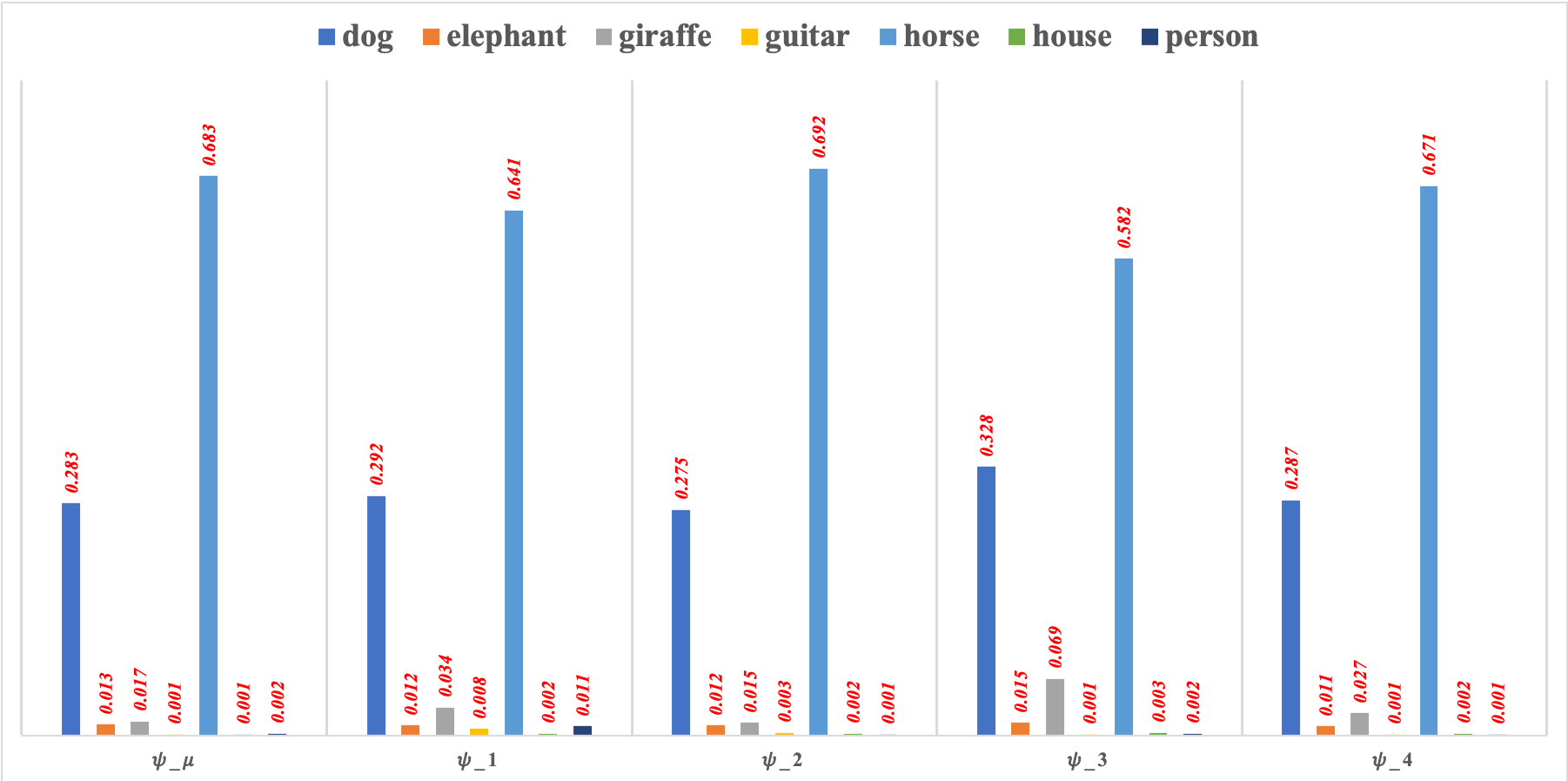}
\caption{The prediction probability of the different sampled classifier $\psi$ for each category (the third image of the success cases in Fig.~\ref{fig:case}).}
\label{fig:c3}
\end{center}
\end{figure*}

\begin{figure*}[htb]
\begin{center}
\includegraphics[width=1.0\textwidth]{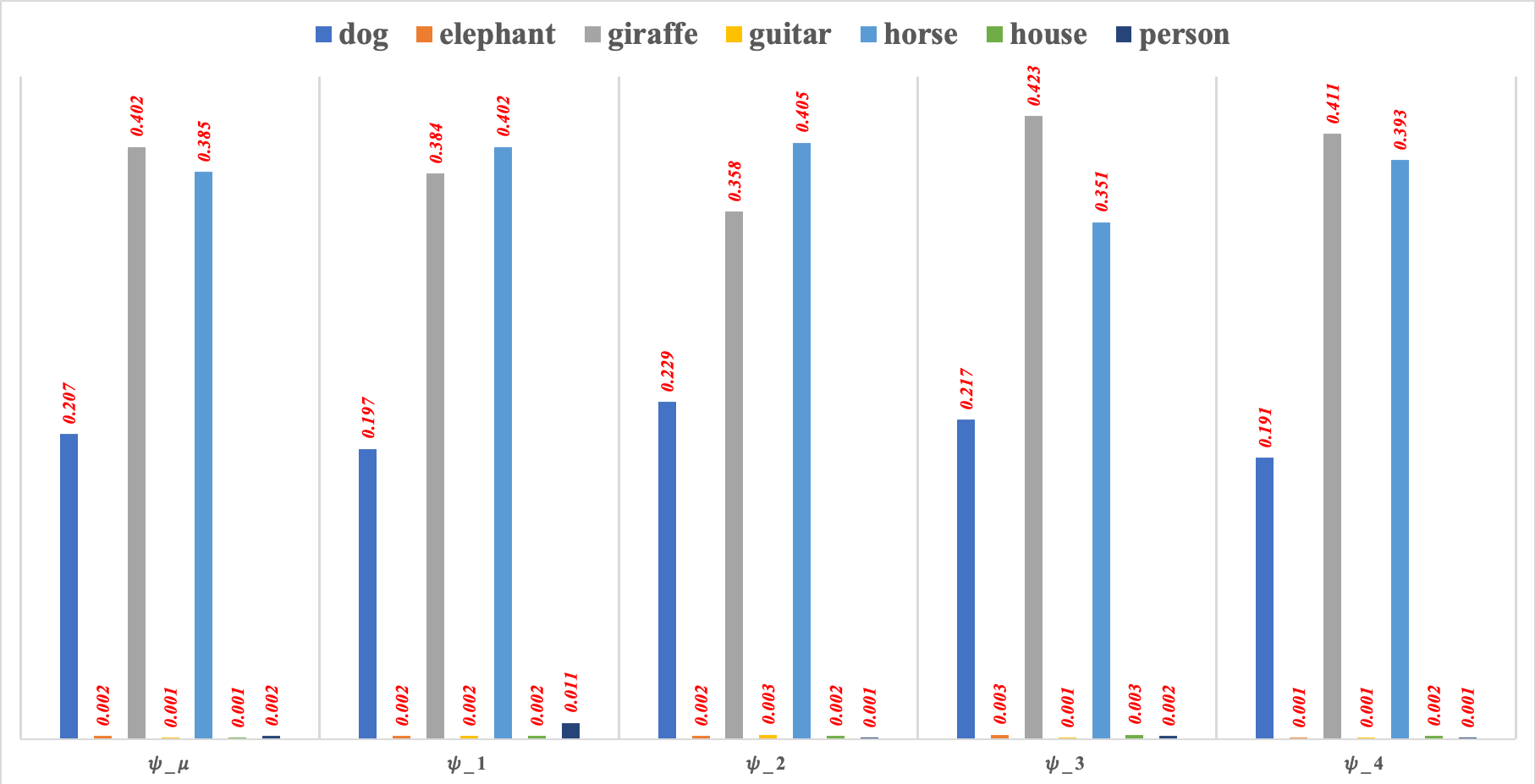}
\caption{The prediction probability of the different sampled classifier $\psi$ for each category (the fourth image of the success cases in Fig.~\ref{fig:case}).}
\label{fig:c4}
\end{center}
\end{figure*}

\begin{figure*}[htb]
\begin{center}
\includegraphics[width=1.0\textwidth]{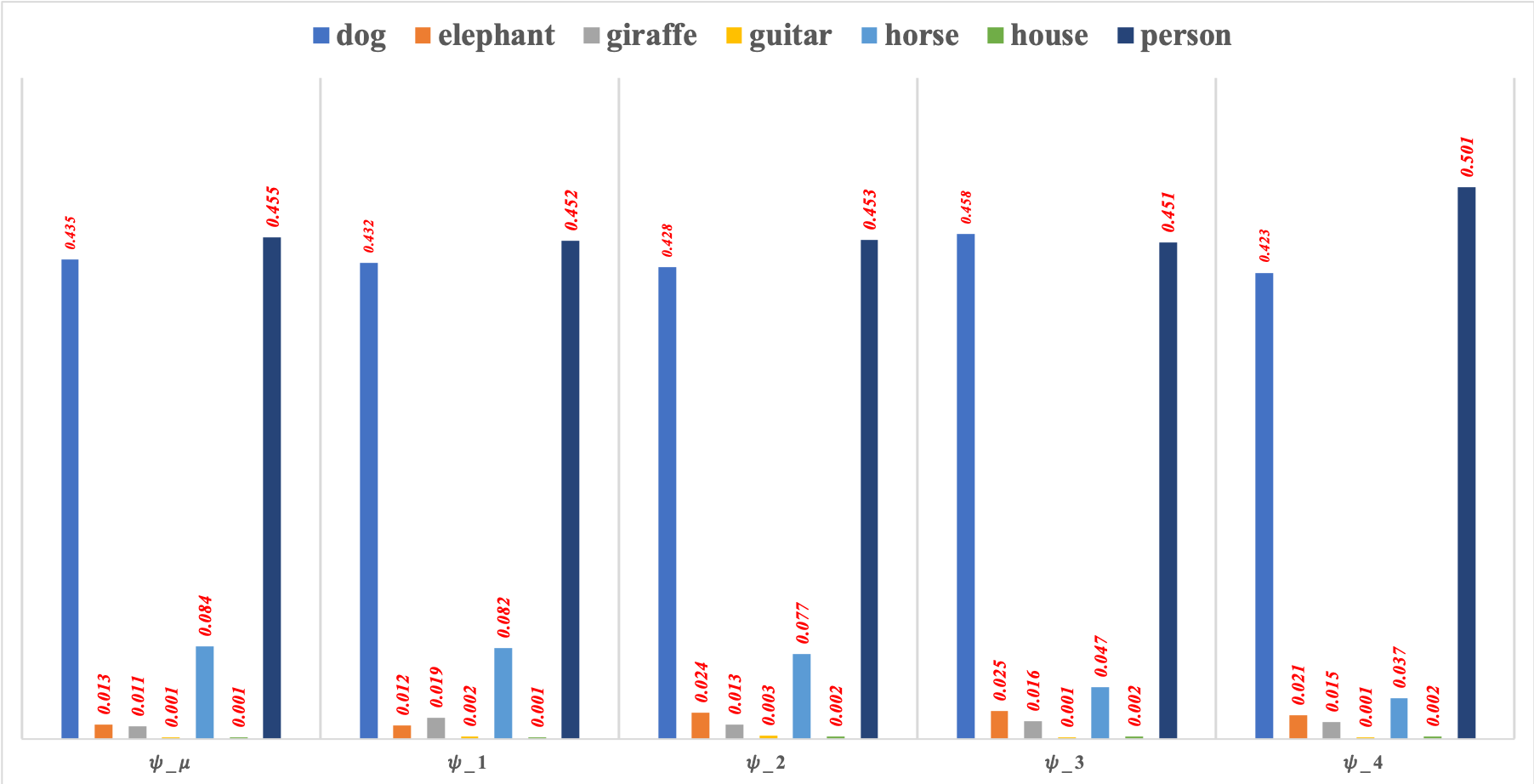}
\caption{The prediction probability of the different sampled classifier $\psi$ for each category (the first image of the failure cases in Fig.~\ref{fig:case}).}
\label{fig:c5}
\end{center}
\end{figure*}

\begin{figure*}[htb]
\begin{center}
\includegraphics[width=1.0\textwidth]{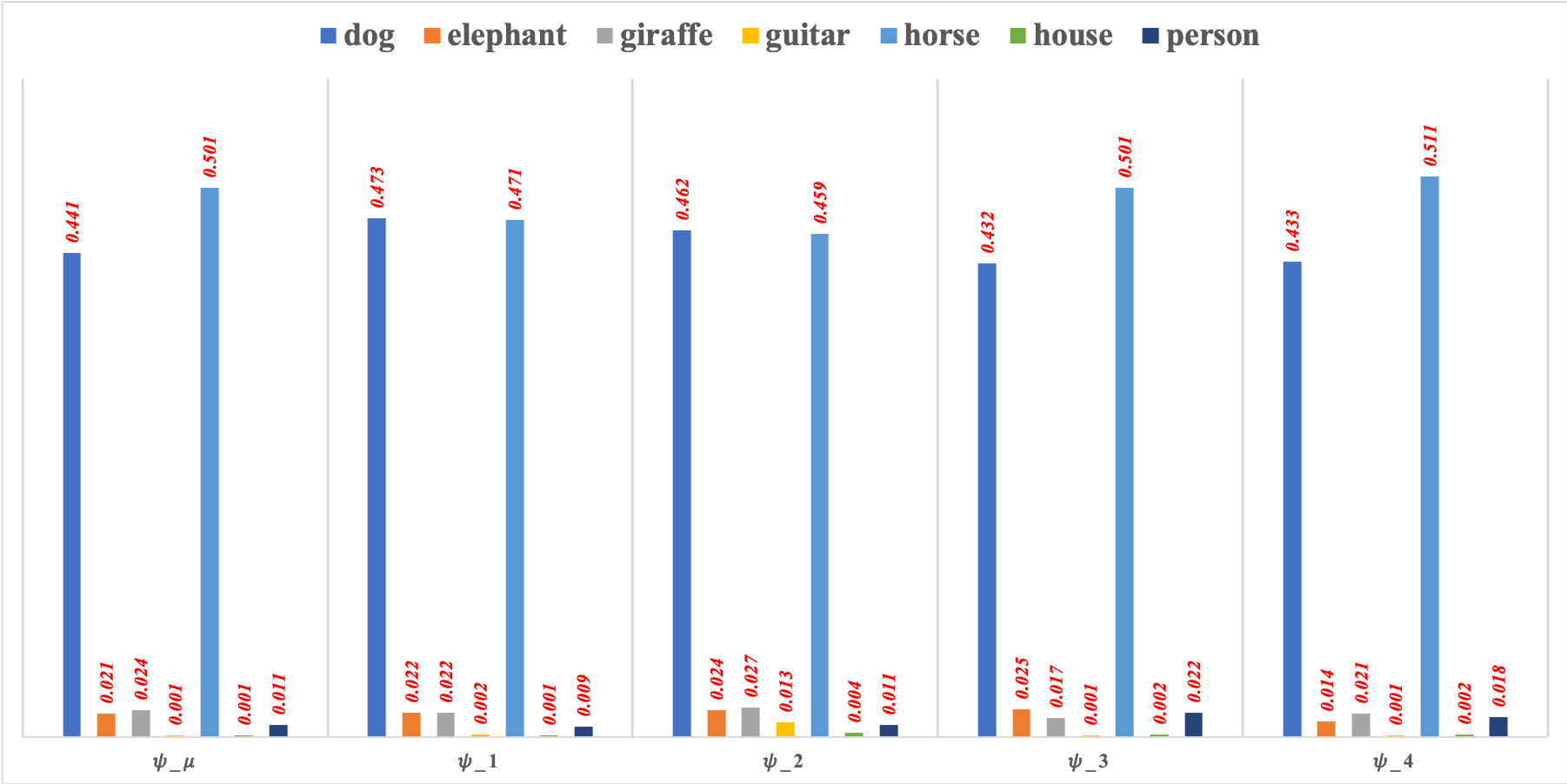}
\caption{The prediction probability of the different sampled classifier $\psi$ for each category(the second image of the failure cases in Fig.~\ref{fig:case}).}
\label{fig:c6}
\end{center}
\end{figure*}

\begin{figure*}[htb]
\begin{center}
\includegraphics[width=1.0\textwidth]{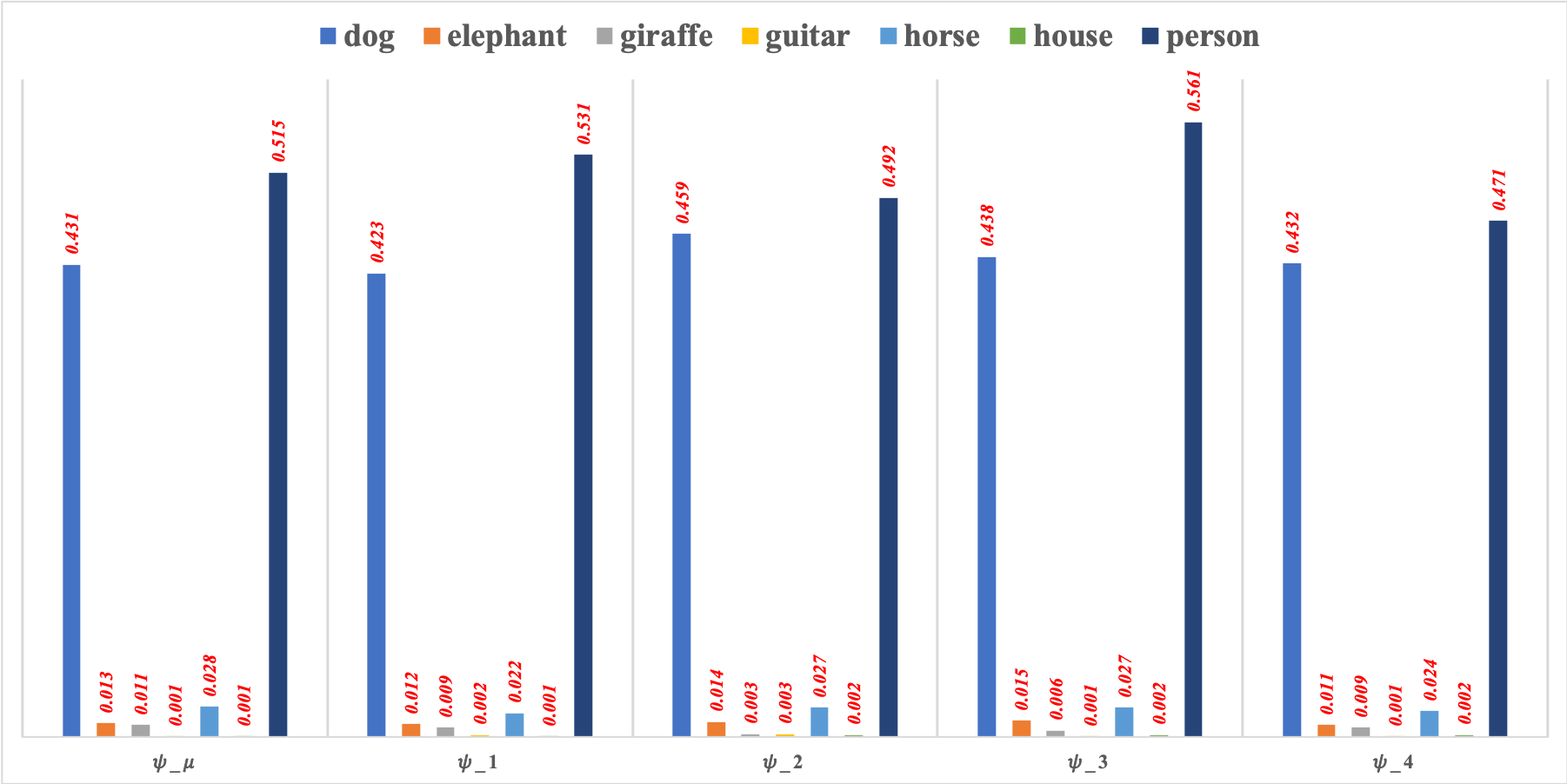}
\caption{The prediction probability of the different sampled classifier $\psi$ for each category(the third image of the failure cases in Fig.~\ref{fig:case}).}
\label{fig:c7}
\end{center}
\end{figure*}

\begin{figure*}[htb]
\begin{center}
\includegraphics[width=1\textwidth]{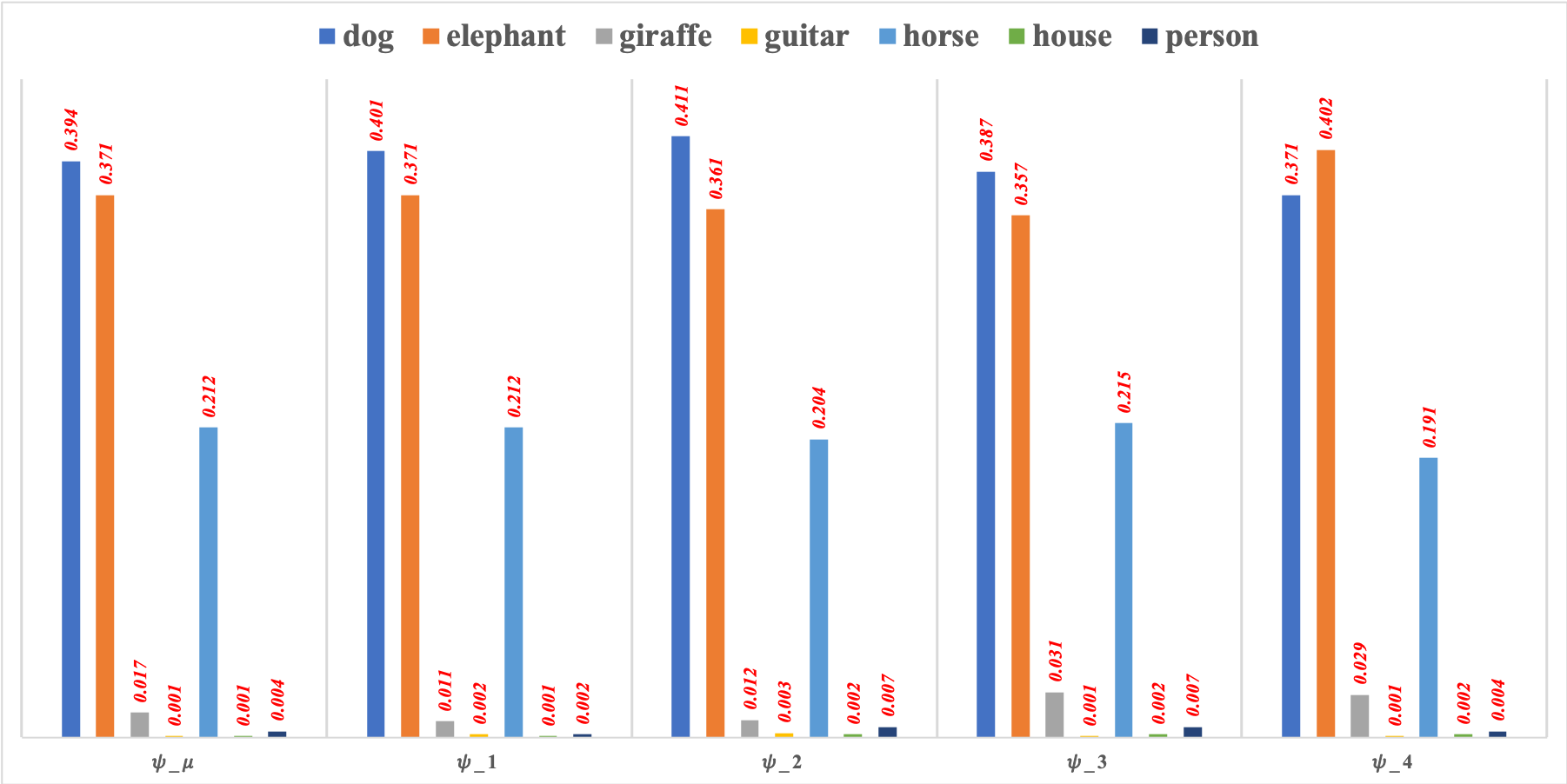}
\caption{The prediction probability of the different sampled classifier $\psi$ for each category (the fourth image of the failure cases in Fig.~\ref{fig:case}).}
\label{fig:c8}
\end{center}
\end{figure*}

In order to demonstrate that the proposed probabilistic modeling can handle prediction uncertainty, we conduct an extra set of experiments as follows:

We shown more  success and failure cases in Fig.~\ref{fig:case} and show the corresponding prediction probabilities of using different sampled classifiers $\psi$ for each category of the image in Fig.~\ref{fig:c1}-\ref{fig:c8}. $\psi$\_$\mu$ indicates the mean value of the classifier. From Fig.~\ref{fig:c1}-\ref{fig:c8}, we can see that different $\psi$ can produce different prediction probabilities to each category. Specially, for the fourth image of success cases, the final result of the classification is \textit{giraffe}. However, the classifiers $\psi$\_$1$ and  $\psi$\_$2$, our model predicts a higher prediction probability of \textit{horse} than \textit{giraffe} as shown in Fig.~\ref{fig:c4}. For the fourth image of failure cases, the image is classified as \textit{dog}, but that the prediction probability of \textit{elephant} is higher than that of \textit{dog} by using classifiers  $\psi$\_$4$ as shown in Fig.~\ref{fig:c8}. Although the final prediction result of our model is incorrect, some of sampled classifiers can still make correct predictions.

\begin{figure*}[htb]
\begin{center}
\subfigure{
\begin{minipage}{0.48\textwidth}
\includegraphics[width=1\textwidth]{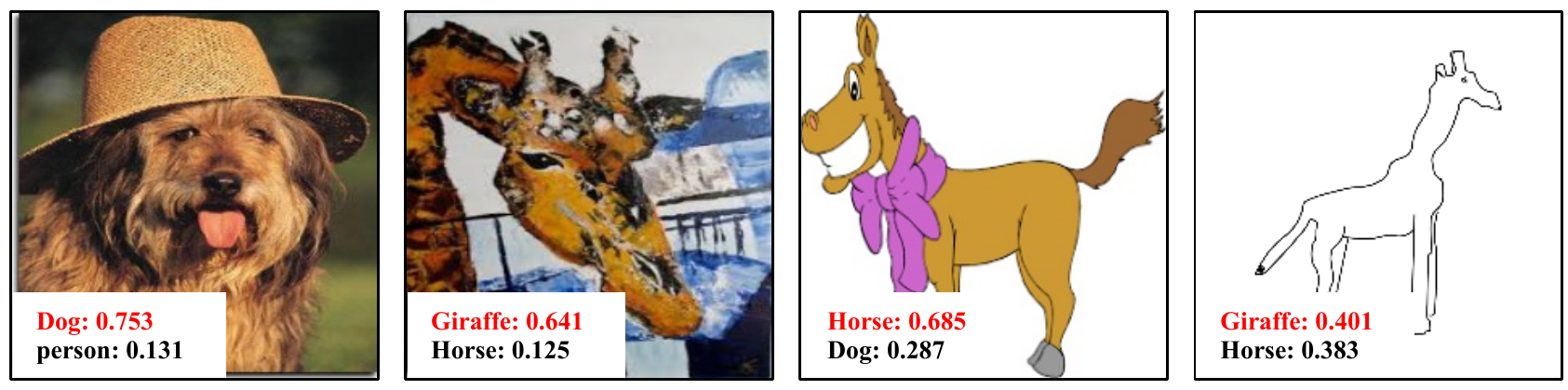}
\centering{(a) Success cases}
\end{minipage}{}
\begin{minipage}{0.48\textwidth}
\includegraphics[width=1\textwidth]{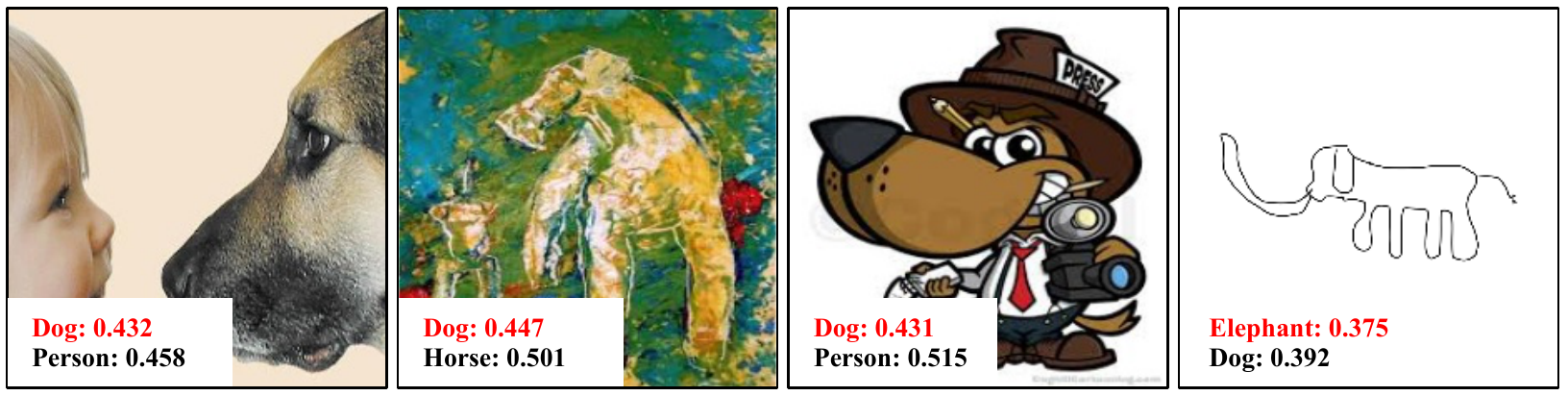}
\centering{(b) Failure cases}
\end{minipage}
}
\caption{\textbf{Success and failure cases} of MetaVIB. The numbers associated with each image are the top two prediction probabilities of MetaVIB, with ground truth labels in red.}
\label{fig:case}
\end{center}
\end{figure*}

\end{document}